\documentclass[10pt,twocolumn,letterpaper]{article}

\usepackage{cvpr}
\usepackage{times}
\usepackage{epsfig}
\usepackage{graphicx}
\usepackage{amsmath}
\usepackage{amssymb}
\usepackage{bm}
\usepackage{enumitem}
\usepackage{booktabs}
\usepackage[normalem]{ulem}

\usepackage[pagebackref=true,breaklinks=true,letterpaper=true,colorlinks,bookmarks=false,pdftitle={Leveraging 2D Data to Learn Textured 3D Mesh Generation}]{hyperref}

 \cvprfinalcopy 

\newcommand{\pmh}[1]{\textcolor{blue}{[PH: #1]}}
\newcommand{\chl}[1]{\textcolor{red}{#1}}

\newcommand{\vt}[1]{\textcolor{red}{[VT: #1]}}

\renewcommand{\pmh}[1]{}
\renewcommand{\chl}[1]{}

\renewcommand{\vt}[1]{}

\newcommand{\sect}[1]{Section~\ref{sec:#1}}
\newcommand{\fig}[1]{Figure~\ref{fig:#1}}
\newcommand{\tab}[1]{Table~\ref{tab:#1}}
\newcommand{\app}[1]{Appendix~\ref{app:#1}}

\makeatletter
\newcommand*{\centerfloat}{%
  \parindent \z@
  \leftskip \z@ \@plus 1fil \@minus \textwidth
  \rightskip\leftskip
  \parfillskip \z@skip}
\makeatother

\newcommand{\masksetting}{(\textsc{mask})}
\newcommand{\bgvaesetting}{(\textsc{no-mask})}
\newcommand{\nimp}{(\textsc{pushing})}
\newcommand{\densep}{(\textsc{dense})}

\newcommand{\zshape}{\ensuremath{\mathbf{z}_\mathrm{shape}}}
\newcommand{\zcolor}{\ensuremath{\mathbf{z}_\mathrm{color}}}
\newcommand{\zbg}{\ensuremath{\mathbf{z}_\mathrm{bg}}}
\newcommand{\pose}{\ensuremath{\bm{\theta}}}

\newcommand{\myparagraph}[1]{\smallskip\noindent\textbf{#1}}

\begin{document}

\title{Leveraging 2D Data to Learn Textured 3D Mesh Generation}

\author{Paul Henderson\\
IST Austria\\
{\tt\small paul@pmh47.net}
\and
Vagia Tsiminaki\\
IBM Research Z\"{u}rich\\
{\tt\small tsi@zurich.ibm.com}
\and
Christoph H. Lampert\\
IST Austria\\
{\tt\small chl@ist.ac.at}
}

\maketitle

\begin{abstract}
    Numerous methods have been proposed for probabilistic generative modelling of 3D objects.
    However, none of these is able to produce \textnormal{textured} objects, which
    renders them of limited use for practical tasks.
    In this work, we present the first generative model of textured 3D meshes.
    Training such a model would traditionally require a large dataset of textured meshes, but unfortunately,
    existing datasets of meshes lack detailed textures.
    We instead propose a new training methodology that allows learning from collections of 2D images
    without any 3D information.
    To do so, we train our model to explain a distribution of images by modelling each image
    as a 3D foreground object placed in front of a 2D background.
    Thus, it learns to generate meshes that when rendered, produce images similar to those in its training set.

    A well-known problem when generating meshes with deep networks is the emergence of self-intersections,
    which are problematic for many use-cases.
    As a second contribution we therefore introduce a new generation process for 3D meshes that
    guarantees no self-intersections arise, based on the physical intuition that faces should push 
    one another out of the way as they move.

    We conduct extensive experiments on our approach, reporting quantitative and qualitative results 
    on both synthetic data and natural images.
    These show our method successfully learns to generate plausible and diverse textured 3D samples
    for five challenging object classes.
    %
\end{abstract}


\setlength{\textfloatsep}{7pt plus 1.0pt minus 2.0pt}
\setlength{\dbltextfloatsep}{7pt plus 1.0pt minus 2.0pt}

\setlength{\abovecaptionskip}{8pt plus 1.0pt minus 2.0pt} 
\setlength{\belowcaptionskip}{8pt plus 1.0pt minus 2.0pt} 

\setlength{\abovedisplayskip}{8pt plus 1.0pt minus 2.0pt}
\setlength{\belowdisplayskip}{8pt plus 1.0pt minus 2.0pt}
\setlength{\abovedisplayshortskip}{8pt plus 1.0pt minus 2.0pt}
\setlength{\belowdisplayshortskip}{8pt plus 1.0pt minus 2.0pt}

\section{Introduction}

Learning the structure of a 3D object class is a fundamental task in computer vision.
It is typically cast as learning a probabilistic generative model, from which instances of the class may be sampled.
%
The last five years have seen dramatic progress on this task~\cite{wu15cvpr-shapenets,wu16nips,rezende16nips,nash17cgf,
li17tog,soltani17cvpr,gadelha173dv,tan18cvpr,xie18cvpr,gadelha18eccv,achlioptas18icml,henderson19ijcv,li19cvpr}, enabled
by new, large-scale training sets~\cite{shapenet15arxiv,wu15cvpr-shapenets}.
However, existing methods generate only shapes, without any associated textures to capture the
surface appearance. 
%
This is a major shortcoming since the surface appearance of an object strongly influences how we perceive and interact
with it---consider for example
the difference between a red and a green tomato, a police car and a taxi, a book and a brick, or a zebra and a horse.
As such, textures are vital for many practical uses of generated shapes, such as visual effects and games.

\begin{figure}
    \centering
    \includegraphics{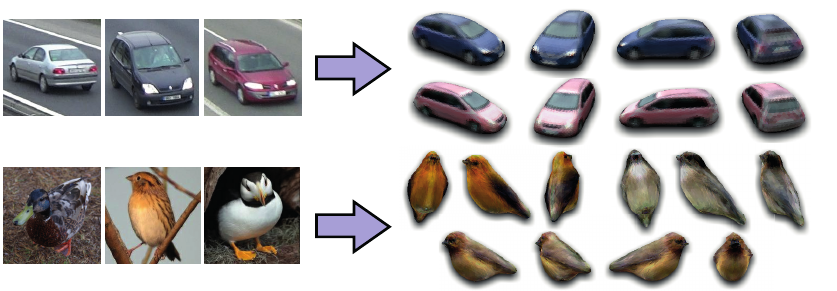}
    \caption{
        We propose a method to learn a generative model of textured 3D shapes (right), from collections of images (left)
    }
    \label{fig:splash}
\end{figure}

We hypothesise that the lack of research on methods that learn to generate textured shapes is in part due to a lack of textured 3D
data for training them.
Of the two large-scale datasets of 3D shapes, ShapeNet~\cite{shapenet15arxiv} lacks detailed textures for most
instances, while ModelNet~\cite{wu15cvpr-shapenets} lacks color information entirely.
%

We propose an alternative paradigm: rather than learning such a model from 3D data, we learn it from a large collection of 2D images (\fig{splash}).
This lets us leverage \textit{existing} weakly-annotated image datasets, at the price of solving a very challenging
learning problem.
It is challenging for three reasons:
(i) infinitely many 3D shapes may project to give the same 2D image;
(ii) we cannot rely on having multiple views of each object instance;
(iii) objects appear in front of cluttered backgrounds, and segmentation masks may not be available.
%
We must therefore learn to isolate foreground objects from background clutter, at the same time as learning their space of valid
shapes and textures.

\begin{figure*}
    \centering
    \includegraphics[width=0.95\linewidth]{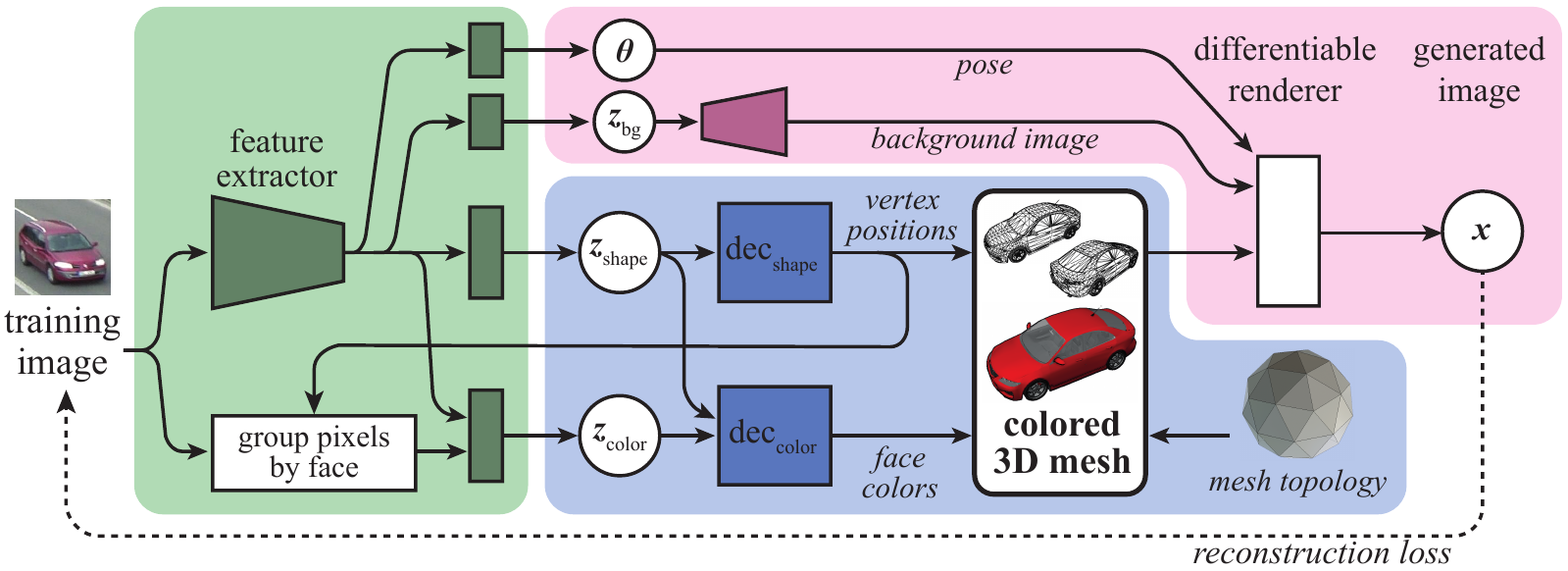}
    \caption{
        We propose a probabilistic generative model of textured 3D meshes (blue; see \sect{mesh-model}).
        We show how to train it using only 2D data (\sect{training}), by adding
        additional components (pink) that model the process of forming an image from a 3D foreground mesh rendered over 2D background clutter.
        We train the model to maximise the likelihood of a dataset of images, by adding an encoder model (green) that predicts the posterior distribution
        on latent variables for a given image.
        White circles represent random variables; colored boxes are densely-connected networks; trapezoids are convolutional networks
    }
    \label{fig:model}
\end{figure*}

Our first contribution is a \textbf{new generative model over textured 3D shapes} (\sect{mesh-model}).
Our second and most significant contribution is a \textbf{method to train this model to match a distribution of images}
(\sect{training}), overcoming the difficulties mentioned.
%
%
Specifically, our model learns to \textit{reconstruct} its training images in terms of the physical process by which they were formed.
This is achieved by augmenting the generative model with an image-formation model---we place the generated 3D object in front of
a camera, and render it over some background to give an image.
An encoder network then predicts the latent parameters that give rise to any given image.
Thus, the model must explain the distribution of training images in terms of a distribution of 3D foreground objects over 2D backgrounds (\fig{model}).
By modelling the variability among object instances using a limited-capacity latent space, we ensure that our method generates
complete, coherent objects,
rather than unrealistic
shapes that could explain each training image in isolation.
Informally, this works because it takes more bits to encode a distribution over many partial, viewpoint-dependent object appearances than
over the variability of the true global appearance model.

We choose meshes as our representation of textured shapes, similar to some recent works on single-image 3D reconstruction~\cite{kanazawa18eccv,
liu19iccv,kato18cvpr,wang18eccv}.
Meshes are the dominant shape representation in computer graphics, and have several benefits over alternatives such as voxels and point-clouds:
(i) their computational cost scales (at worst) with surface area not volume; (ii) they can represent arbitrarily-oriented surfaces;
(iii) they can directly represent solid surfaces with a well-defined normal and interior/exterior.

Training a model to output meshes that correctly reconstruct its training images suffers one potential failure mode---correct-looking
images may be obtained by rendering even highly-irregular, self-intersecting meshes, due to the ambiguity of the projection operation.
This is problematic as many downstream use-cases, such as physical simulations, geometry-processing algorithms, and 3D printing,
require meshes that are \textit{non-intersecting}---that is, no triangular faces intersect with any others.
For smooth, regular object classes such as cars, non-intersection can be encouraged by careful regularization of the
local surface geometry~\cite{kato18cvpr,kanazawa18eccv}.
%
However, for angular, non-convex object classes with elongated parts, such as chairs and airplanes, a sufficiently
strong regularizer results in overly-smoothed surfaces lacking in detail.

As a further contribution, we therefore propose a \textbf{novel mesh parametrization, that \textit{necessarily} yields non-intersecting
surfaces even without regularization} (\sect{non-intersection}).
It nonetheless has great representational flexibility, and can faithfully capture complex shapes such as chairs and airplanes.
Our parametrization has a simple physical intuition---as faces move, they push others out of the way rather than intersecting
them; we show how to formalise this idea and efficiently incorporate it in gradient-based training.

%
We conduct extensive experiments illustrating the performance of our method (\sect{experiments}).
We first validate its performance on synthetic data from four diverse object classes, then show that it also performs well on two
challenging classes of natural images.
In all cases, we show both quantitatively and qualitatively that our method successfully learns to generate samples that
are diverse and realistic, even when ground-truth segmentation masks are not available.
%
Moreover, we show that our novel mesh parametrization eliminates problematic self-intersections, yet allows representing
angular and concave classes such as chairs, airplanes, and sofas.
%


\section{Related work}
\label{sec:rel-work}

The vast majority of methods for automatically generating 3D shapes are based on variational autoencoders
(VAEs)~\cite{kingma14iclr} or generative adversarial networks (GANs)~\cite{goodfellow14nips}.
However, almost all such models (\eg \cite{wu15cvpr-shapenets,xie18cvpr,tan18cvpr,zhu18nips,balashova183dv,
gadelha18eccv,achlioptas18icml}) must learn from large datasets of 3D shapes~\cite{shapenet15arxiv,wu15cvpr-shapenets};
this stands in contrast to our own, which instead learns from collections of images.
We now discuss existing works that do learn (untextured) 3D generative models from 2D datasets, and
also methods for single-image reconstruction that are trained with only 2D data.
%
%
%

\myparagraph{Generative 3D models learnt from 2D data.}
The first work to learn a generative model of 3D shapes from 2D data was \cite{gadelha173dv}.
The authors train a GAN that produces voxels, with a discriminator that ensures these project to silhouettes matching a
distribution of ground-truth segmentation masks.
MP-GAN~\cite{li19cvpr} extends this to use multiple discriminators, corresponding to different viewpoints, simultaneously
training a classifier to predict the viewpoint.
Reliance on silhouettes is a severe limitation when ground-truth masks are not available; our method avoids it
by reconstructing the image pixels themselves.
The closest work in spirit to ours is \cite{henderson19ijcv}, which learns single-image reconstruction and mesh generation
from untextured renderings under known lighting.
Unlike us, they do not to learn to generate or reconstruct textures, so their method still cannot work on natural images.
Moreover, their method has no mechanism to ensure that it produces meshes without self-intersections.
Finally, HoloGAN~\cite{nguyenphuoc19iccv} is a GAN over images, that incorporates a 3D latent feature space.
This allows manipulation of 3D pose parameters somewhat-independently of object identity and background.
However, it cannot output an explicit 3D shape representation such as a voxel grid or mesh.
%
%
%

\myparagraph{Single-image reconstruction learnt from silhouettes.}
Several recent works learn single-image 3D reconstruction from 2D silhouettes~\cite{rezende16nips,yan16nips,wiles17bmvc,
tulsiani17cvpr, kato18cvpr,tulsiani18cvpr-mvc,kanazawa18eccv,insafutdinov18nips,yang18eccv,henderson18bmvc,henderson19ijcv,liu19iccv,kato19cvpr,henzler19iccv}.
These are trained discriminatively to map images to 3D representations (voxels, meshes, or point-clouds); they use
losses that ensure the reprojected, reconstructed silhouette matches one or more masks provided as supervision.
Four of these works also consider colors:
\cite{liu19iccv} has a post-processing stage that predicts the texture for a reconstructed mesh;
\cite{tulsiani17cvpr} shows an example of reconstructing colored voxels, assuming training with multiple views per instance;
\cite{henzler19iccv} does the same even when only a single view per instance is available;
\cite{kanazawa18eccv} learns single-image textured mesh reconstruction, using mask
and keypoint annotations.
%
Unlike ours, none of these methods allow sampling new meshes \textit{a priori}---they do not learn an explicit prior.

\myparagraph{Texture generation and reconstruction.}
Other recent works learn to generate or reconstruct textures given full supervision, which limits them to
classes for which extensive textured 3D data is available.
\cite{oechsle19iccv} defines textures implicitly, as a function of 3D position; they show that this representation allows sampling
textures given a 3D shape, or reconstructing texture from a single image and shape.
\cite{zhu18nips} generates images of textured 3D shapes, by first sampling textureless shapes and projecting them to a silhouette
and depth-map; the texture is then generated in image space.
\cite{sun18arxiv} performs textured single-image reconstruction, using colored voxels as the output representation.
\cite{li19cvpr-3dasr,richard193dv} learn priors on textures to improve multi-image reconstruction.
%

\myparagraph{Non-intersecting mesh parametrization.}
One of our contributions is a method to parametrize meshes such that the resulting surface is highly flexible, yet guaranteed
not to intersect itself.
Methods for single-image 3D reconstruction use local smoothness regularizers such as Laplacian regularization~\cite{sorkine05eg}, TV-L1
regularization~\cite{zhang15tvcg}, and regularizing the angles between adjacent faces~\cite{kato18cvpr,liu19iccv}.
However, these only prevent local surface irregularity, and do not penalise two smooth surfaces passing through each other.
The related problem of detecting, characterising and removing mesh self-intersections has received some attention in the graphics
literature~\cite{yamakawa09pimrt,campen10eg,jacobson13tog}.
Unfortunately, none of these methods allow us to construct meshes that do not intersect \textit{a priori}, which is necessary when
we predict their shape directly from a neural network.


\section{Modelling textured meshes}
\label{sec:mesh-model}


We begin by defining our probabilistic generative model for textured 3D meshes (\fig{model}, blue background).
Each mesh consists of $N_V$ vertices, and $N_F$ triangular faces to which we assign colour values $\mathbf{c}$.
We assume fixed topology for all meshes, \ie $N_V$, $N_F$, and the mapping between faces and vertices, do not vary between instances.
To generate a mesh, the model must therefore sample the positions $\mathbf{v}$ of all vertices, and the colors $\mathbf{c}$ of all faces.

We draw low-dimensional latent code variables from standard Gaussian distributions, then pass these through
\textit{decoder} networks producing the required attributes:
%
%
\begin{align}
    \zshape &\sim \mathrm{Normal}(\mathbf{0}, \, \mathbf{1}) \\
    \zcolor &\sim \mathrm{Normal}(\mathbf{0}, \, \mathbf{1}) \\
    \mathbf{v} &= \mathrm{dec}_\mathrm{shape} \left( \zshape \right) \\
    \mathbf{c} &= \mathrm{dec}_\mathrm{color} \left( \zcolor, \, \zshape \right)
\end{align}
Here, \zshape{} captures the object's 3D shape; the shape decoder $\mathrm{dec}_\mathrm{shape}$ can be as simple as a
densely-connected ELU network with $3 N_V$ outputs $\mathbf{v}$, corresponding to the 3D position of each vertex (we discuss a more sophisticated option later).
Similarly, \zcolor{} captures texture; $\mathrm{dec}_\mathrm{color}$ is a densely-connected ELU network with $3 N_F$ outputs $\mathbf{c}$,
corresponding to the color of each face represented as red/green/blue values.

Note that separating \zshape{} from \zcolor{} lets us reconstruct the shape of an instance
before its color is known, which will be important for our training process (\sect{training}).
However, by passing the shape code into $\mathrm{dec}_\mathrm{color}$,  we still allow dependencies between color and shape,
\eg to capture the fact that a bird with the shape of an eagle should never have the colors of a robin.
Detailed network architectures are given in \app{architectures}.

Calculating vertex locations with a neural network typically results in highly irregular meshes with many self-intersections (\eg \fig{pushing-vs-regular-examples}).
This is undesirable for many use-cases, and in general, difficult to avoid with regularization alone.
The next section describes a more sophisticated structure for $\mathrm{dec}_\mathrm{shape}$, that produces vertex locations which
are guaranteed not to create intersecting surfaces.

\section{Non-intersecting mesh parametrization}
\label{sec:non-intersection}

Our goal is a decoder network that produces vertex locations such that no surface intersections can occur, but
highly concave, angular surfaces (\eg chairs) can still be represented.

\myparagraph{Physical motivation.}
%
When playing with a deflated balloon,
we can deform it quite
drastically without introducing any intersections, simply because when attempting to push one surface through another,
the second will instead be pushed out of the way.
It is not computationally feasible to simulate the physical dynamics of such a system during training.
Nonetheless, we can make use of this insight, by combining a careful choice of parametrization with a simple, efficient model of surface collisions.


\newcommand{\dbefore}{\ensuremath{\mathbf{d}^\mathrm{min}}}
\newcommand{\dbeforev}{\ensuremath{d^\mathrm{min}_v}}
\newcommand{\dafter}{\ensuremath{\mathbf{d}}}
\newcommand{\dafterv}{\ensuremath{d_v}}
\newcommand{\direction}{\ensuremath{\hat{\mathbf{u}}}}

\myparagraph{Parametrization.}
Instead of producing the final set of vertex positions in a single shot, we perform a sequence of $N_s$ simpler deformation
steps, starting from an initial, non-learnt mesh with spherical topology (and thus no intersections).
In each step, we will move all vertices in the same direction \direction{}, but by differing (non-negative) distances \dafter{}.
A densely-connected network outputs \direction, and \textit{lower bounds} \dbefore{} on the distances,
but these are modified to give \dafter{} such that each face pushes others rather than intersecting them.

\begin{figure}
    \centering
    \includegraphics[width=0.95\linewidth]{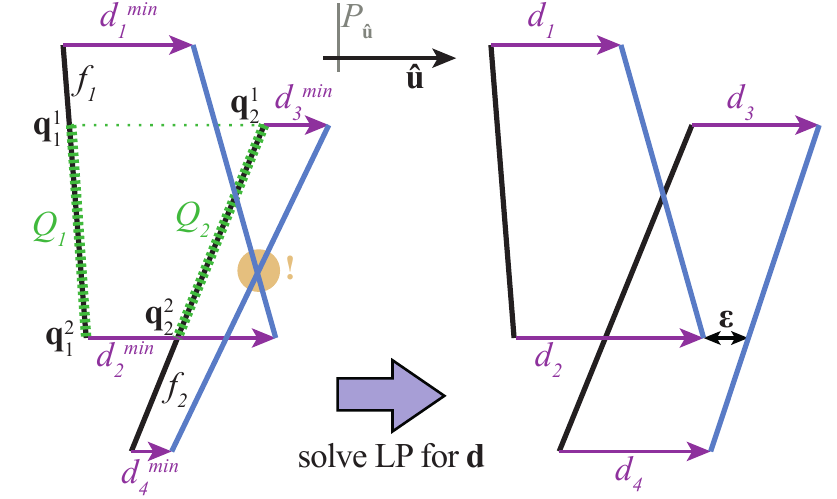}
    \caption{
        A 2D example of pushing faces (left is before pushing, right is after); \direction{} lies in the plane of the page, with
        $P_{\direction{}}$ perpendicular.
        Black lines are the initial faces $f_1, f_2$; blue lines are after shifting by \dbefore{} (left) and \dafter{} (right).
        $f_1, f_2$ do not initially intersect, but overlap in $P_{\direction{}}$; the proposed distances \dbefore{} of motion
        along \direction{} would create an intersection in the orange circle.
        \dbefore{} is mapped to \dafter{} such that the intersection is prevented, by solving an LP; note that $\dafterv > \dbeforev$
        for $v \in \{3, \, 4\}$, but $\dafterv = \dbeforev$ for $v \in \{1, \, 2\}$.
    }
    \label{fig:pushing}
\end{figure}

\myparagraph{Pushing faces.}
%
We map the initial distances \dbefore{} to pushed ones \dafter{} that minimise $\sum_v \dafterv$, where $v$ indexes vertices.
This minimisation is subject to two sets of constraints:
(i) $\dafterv \geq \dbeforev \; \forall v$, \ie each vertex moves at least as far as
specified by the decoder network, and
(ii) no intersections should arise.
%
%

To impose (ii),
we first find all pairs of faces that \textit{could} intersect, depending on the distances their vertices move.
As all vertices move in the same direction, this problem can be considered in 2D: we project all vertices and faces
into a plane $P_{\direction}$ perpendicular to the direction of motion \direction{}, and find all pairs of faces for which the corresponding
triangles in $P_{\direction}$ intersect with non-zero area.
For each such pair of faces $(f_1, \, f_2)$, their intersection is a polygonal region of $P_{\direction}$; we can re-project this region back onto
each face in 3D space, giving planar polygons $(Q_1, \, Q_2)$ respectively (\fig{pushing}).

As there are initially no intersections, we can order $Q_1$ and $Q_2$ according to their projection onto \direction{}:
if $\mathbf{q}_i^j$ is the position of the $j$\textsuperscript{th} corner of $Q_i$, then either
$\mathbf{q}_1^j \cdot \direction{} < \mathbf{q}_2^j \cdot \direction{} \; \forall j$, or
$\mathbf{q}_1^j \cdot \direction{} > \mathbf{q}_2^j \cdot \direction{} \; \forall j$.
We assume without loss of generality that the former holds;
to avoid intersections, this must remain true even after moving each vertex by $\dafterv \direction$.

Let $\bm{\beta}_i^j$ be the barycentric coordinates of
$\mathbf{q}_i^j$
\wrt the triangular face $f_i$ in which it lies.
The distance by which $\mathbf{q}_i^j$ moves is then given by $\bm{\beta}_i^j \cdot \dafter_{\,|f_i}$, where
$\dafter_{\,|f_i}$ contains the elements of \dafter{} corresponding to the vertices of face $f_i$.
%
%
The constraint that $f_1$ and $f_2$ do not intersect, \ie that none of the points $\mathbf{q}_i^j$ change ordering,
then becomes
\begin{equation}
    \mathbf{q}_1^j \cdot \direction{}
    +
    \bm{\beta}_1^j \cdot \dafter_{\,|f_1}
    + \varepsilon
    \leq
    \mathbf{q}_2^j \cdot \direction{}
    +
    \bm{\beta}_2^j \cdot \dafter_{\,|f_2}
    \; \,
    \forall j
\end{equation}
where $\varepsilon$ is a small buffer distance.

All the constraints we have defined are linear in \dafter{}.
Thus, minimising our objective $\sum_v \dafterv$ under them defines a linear programming problem (LP),
which can be solved efficiently using the simplex algorithm~\cite{Nocedal06}.
In practice, we use the efficient off-the-shelf solver of \cite{gurobi19}.

\myparagraph{Propagating derivatives.}
We have now defined \dafter{} as the solution to an optimisation problem that has \dbefore{} and the vertex locations
as inputs.
In order to incorporate this in our decoder network, we need to propagate gradients back through the optimisation
process from \dafter{} to these inputs.
Note that the solution to an LP always lies at a corner of the polytope defined by its constraints.
At this corner, equality holds for some subset of \textit{active} constraints.
These constraints define a system of linear equations, whose solution equals \dafter{}.
Thus, back-propagating gradients from \dafter{} is exactly equivalent to back-propagating them through the
process of solving this linear system, \eg by Cholesky decomposition, allowing direct implementation in
TensorFlow~\cite{abadi15tensorflow}.
%

\section{Training from images}
\label{sec:training}


Our goal is to train the generative model of \sect{mesh-model} using only images, without any 3D data
We assume access to a training set of images, each containing exactly one instance of the target object class, and
consider two training settings:
\begin{itemize}
    \item{
        \masksetting~~
        We have access to (i) the approximate camera calibration; (ii) a segmentation mask for each target object instance;
        and (iii) the background image, \ie a view of the same environment but without the foreground object present.
        For our experiments, we estimate these automatically from weakly-annotated data, \eg by running a
        segmentation algorithm on unannotated data and inpainting the background, and estimating the
        camera calibration from keypoints.
        We found the model is robust to even quite large errors in the camera calibration.
        This setting is similar to some weakly-supervised methods for single-image 3D reconstruction~\cite{kanazawa18eccv,kato19cvpr}
        and untextured 3D generation~\cite{gadelha173dv,henderson19ijcv}.
    }
    \item{
        \bgvaesetting~~
        We have only the (approximate) camera calibration available.
        This second setting is much more challenging, and goes beyond all prior works on weakly-supervised reconstruction
        and generation.
    }
\end{itemize}

To allow training in these settings, we augment the generative model with additional components to model the
entire image formation process (\fig{model}, pink background).
Specifically, after sampling the mesh itself, we position it in 3D space in front of a perspective camera, and render it over a background image.
The final, observed image $\mathbf{x}$ is an isotropic Gaussian random variable with mean equal to the rendered pixels, and fixed variance.
We then introduce an encoder (or inference) network, that predicts the latent variables corresponding to a given image.
This lets us train our model to match a distribution of \textit{images} (rather than meshes), by learning to reconstruct each in
terms of a foreground mesh in front of background clutter.
%
%
%
We now describe each aspect of this training methodology in detail.


%
%
%


\myparagraph{Background image.}
In setting \masksetting, the background is provided as input to the model along with the training image.
In setting \bgvaesetting, we explicitly model the background, \ie our generative process samples both the 3D foreground
object, and the 2D pixels that are `behind' it.
The background is generated by sampling a low-dimensional latent code vector \zbg{}, and passing
it through a convolutional decoder network.
%
Whereas the decoder in a VAE or GAN is typically designed to be as powerful as possible,
we need to avoid the background model being \textit{too powerful}, and thus able to model the foreground object as well.
%
We therefore set the dimensionality of \zbg{} to be just 16, use only three transpose-convolutional layers, and
upsample the resulting image $4 \times$ or $6 \times$ to the desired resolution.
%
This ensures the model cannot capture high-frequency details such as the edge of the foreground object.

%
%

\myparagraph{Rendering.}
To render the generated mesh over the background image, we first place it in 3D space relative to a camera at the origin, according to a
pose $\bm{\theta}$.
This captures the fact that the object may not be centered nor at known, constant depth in the original image, and we do not wish the
shape model to have to account for this variability.
We then project and rasterise the mesh, using the publicly-available differentiable mesh renderer \textit{DIRT}~ \cite{henderson18github,henderson19ijcv}.
%
%
We use direct illumination and Lambertian reflectance~\cite{lambert60bk}, and do not attempt to disentangle albedo from shading.


\myparagraph{Variational training.}
Let $\mathbf{z}$ denote all the latent variables $(\zshape, \, \zcolor, \, \zbg, \, \bm{\theta})$.
Our full model defines a joint distribution $P(\mathbf{x}, \, \mathbf{z}) = P(\mathbf{z}) P(\mathbf{x} \, | \, \mathbf{z})$ over these and the resulting
image $\mathbf{x}$.
We would ideally train it to maximise the likelihood of a training dataset of images; however, this is intractable due to the latent variables.
We therefore adopt a variational approach~\cite{Jordan99}, adding an encoder network that predicts parameters of a variational
posterior distribution $Q \left( \mathbf{z} \,| \, \mathbf{x} \right)$.
%
The following is then a lower bound (the \textit{ELBO}) on the data log-likelihood~\cite{rezende14icml,kingma14iclr}:
\begin{equation}
    \label{eq:loss}
    \mathcal{L} =
    \mathop{\mbox{\large$\mathbb{E}$}}_{Q(\mathbf{z} | \mathbf{x})} \log P(\mathbf{x} | \mathbf{z})
    \, - \, D_{\mathrm{KL}} \left[ Q(\mathbf{z} | \mathbf{x}) \, || \, P(\mathbf{z}) \right]
    \leq
    \log P(\mathbf{x}).
\end{equation}
In practice, rather than maximising $\mathcal{L}$ directly, we make two modifications.
First, following \cite{higgins17iclr}, we multiply the KL-divergence term by a constant factor.
Second, we replace the Gaussian log-likelihood $\log P(\mathbf{x} | \mathbf{z})$ by the multi-scale structural similarity
(MS-SSIM) metric~\cite{wang03acssc}, which (i) allows gradients to propagate across longer distances in the image, and (ii) tends to preserve fine details better.
%
The generative model and encoder network are trained jointly to maximise this objective---thus they learn to reconstruct input images, while ensuring
the posterior distribution on the latents is close to the prior.

\begin{figure}
    \centering
    \includegraphics[width=\linewidth,trim=0 20 0 26,clip]{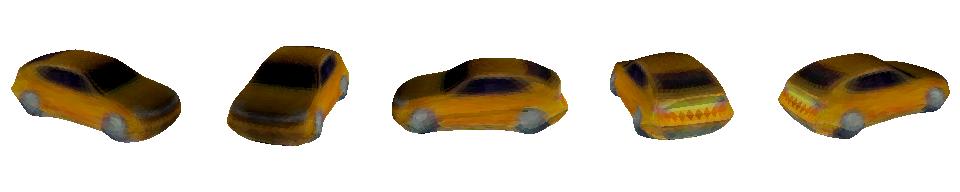}\\
    \includegraphics[width=\linewidth,trim=0 20 0 26,clip]{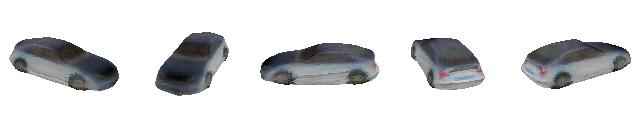}\\
    \includegraphics[width=\linewidth,trim=0 20 0 26,clip]{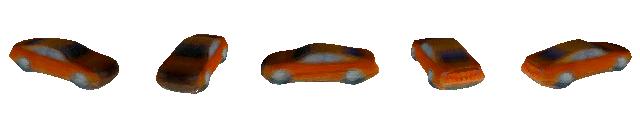}\\
    \includegraphics[width=\linewidth]{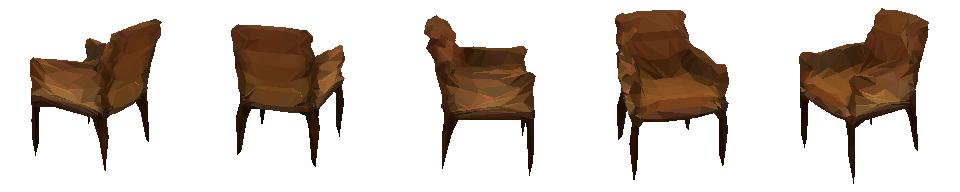}\\
    \includegraphics[width=\linewidth]{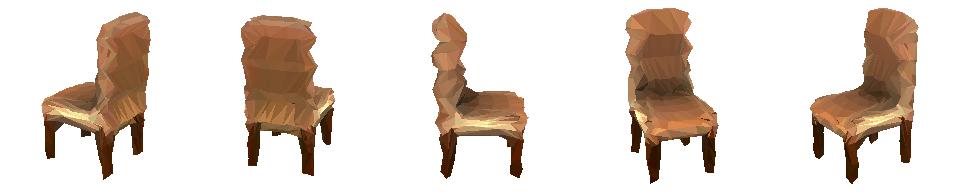}\\
    \includegraphics[width=\linewidth]{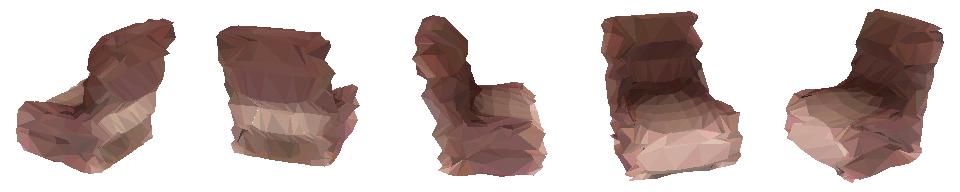}\\
    \includegraphics[width=\linewidth,trim=0 50 0 50,clip]{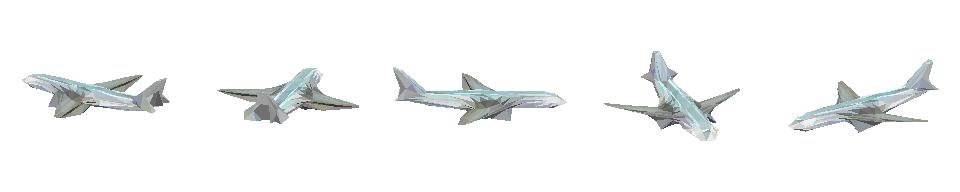}\\
    \includegraphics[width=\linewidth,trim=0 50 0 50,clip]{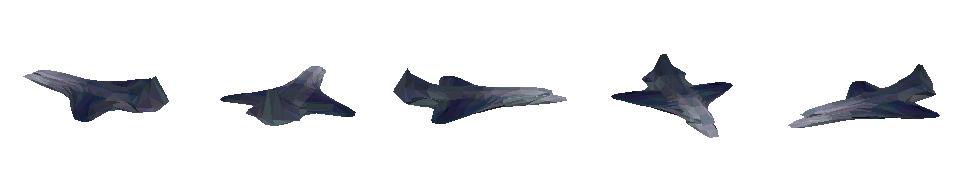}\\
    \includegraphics[width=\linewidth,trim=0 50 0 50,clip]{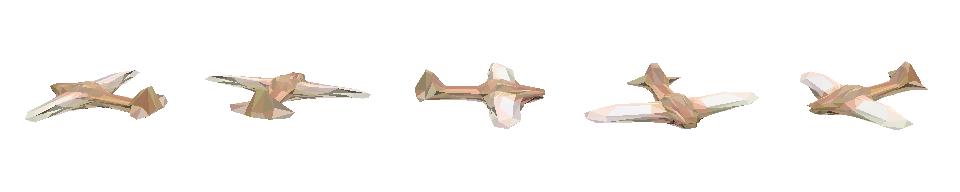}\\
    \includegraphics[width=\linewidth,trim=0 20 0 30,clip]{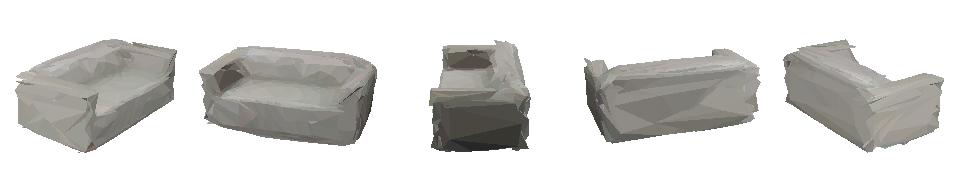}\\
    \includegraphics[width=\linewidth,trim=0 20 0 30,clip]{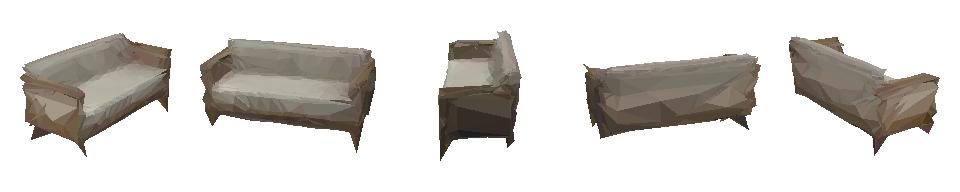}\\
    \includegraphics[width=\linewidth,trim=0 20 0 30,clip]{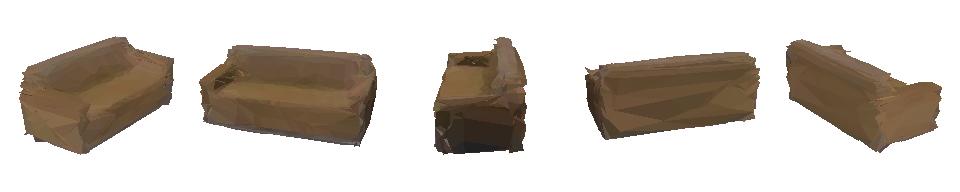}
    \caption{
        Textured meshes sampled from our model, trained on renderings from ShapeNet.
        Each row of five images shows the same sampled mesh from different viewpoints.
        Cars and sofas are trained in setting \bgvaesetting{} with \densep{} parametrization; chairs and airplanes are trained
        in setting \masksetting{} with \nimp{} parametrization, and thus are free of self-intersections in spite of the finely
        detailed geometry.
        Note the samples are diverse and realistic, both in terms of the mesh geometry and textures.
    }
    \label{fig:class-vs-setting-examples}
\end{figure}

\myparagraph{Encoder.}
The encoder network (\fig{model}, green background) takes an image as input, and predicts the mean and variance of Gaussian posterior
distributions on \zshape{}, \zcolor{}, \zbg{}, and \pose{}.
We use a small CNN (similar to \cite{wiles17bmvc,henderson19ijcv}) to extract features from the image; 
the mean and log-variance of \zshape{}, \zbg{}, and \pose{} are then computed by densely-connected layers.
Detailed network architectures are given in \app{architectures}.

We experimented with the same approach for \zcolor{}.
However, this gives rise to a challenging learning task: the network must map an image to the colors of all faces,
without explicitly knowing which faces would be visible in the reconstruction, nor where in the image they project to.
We achieved better results using a novel architecture that explicitly incorporates this information.
We group pixels of the input image (\fig{model}, `group pixels by face') according to which (if any) face of the reconstructed mesh defined by \zshape{} they would
lie in (this is possible as the shape does not depend on \zcolor{}, hence can be reconstructed first; see \sect{mesh-model}).
For each group (including one for background pixels), we calculate the mean RGB values of the pixels that were assigned to it.
This defines a matrix of size $(N_F + 1) \times 3$, which we flatten and encode with a small densely-connected network.
The resulting code then captures the input pixel colors in a way that accounts for where
they lie with respect to the surface of the reconstructed mesh.
Thus, the encoder network does not have to learn this invariance
itself from data.
Finally, the mean and variance for \zcolor{} are estimated by another dense layer taking this code and the image features as input.


\myparagraph{Regularization.}
In setting \masksetting, we additionally maximise the intersection-over-union (IOU) between the ground-truth mask and
the reconstructed silhouette.
This ensures the foreground mesh does not overlap into background regions of the image.
For easier optimisation, we take the mean of the  IOU at multiple scales, using smoothed and downsampled masks/silhouettes.
%
We also found it useful to add local mesh smoothness regularizers, similar to \cite{kanazawa18eccv,kato18cvpr,liu19iccv}.
%

\section{Experiments}
\label{sec:experiments}

We conduct experiments on five diverse object classes: birds, cars, airplanes, chairs, and sofas, which have also
been the focus of related works on weakly-supervised reconstruction and generation~\cite{kanazawa18eccv,li19cvpr,henderson19ijcv,tulsiani18cvpr-mvc}.
%
In \sect{experiments-shapenet}, we validate our method in a controlled setting on renderings of ShapeNet
meshes~\cite{shapenet15arxiv}, analysing its performance under different settings and parametrizations.
Then, in \sect{experiments-real}, we show that it successfully learns models from two challenging datasets
of natural images.
Finally, in \sect{experiments-recon}, we show that the trained model can also be used for single-image 3D
reconstruction on natural images.

All hyperparameters (\eg regularizer strengths and mesh resolution) were selected manually
for perceived quality of generation; we did not directly optimise them \wrt our quantitative metrics.
Our code, hyperparameters and preprocessed datasets will be made available online very soon.


\myparagraph{Metrics.}
%
%
We are not aware of any existing evaluation metric for generative models of textured 3D shapes.
%
%
%
%
We therefore propose a new evaluation protocol using established 2D metrics: inception score (IS)~\cite{salimans16nips},
Fr\'{e}chet inception distance (FID)~\cite{heusel17nips} and kernel inception distance (KID)~\cite{binkowski18iclr}.
All these metrics pass a large set of generated images through the CNN of \cite{szegedy16cvpr}, recording output
logits and feature activations.
IS measures how similar the logits are to a uniform distribution, in terms of KL divergence (larger is better).
FID and KID pass a test set of ground-truth images through the same CNN, then measure how similar their feature
activations are to those of the generated images (smaller is better).
%
%
To apply these metrics in our setting, we render 25600 meshes sampled from our model, each over a ground-truth background, and
report the IS/FID/KID values for these images~\footnote{use of ground-truth backgrounds avoids confounding the quality of mesh
generation with that of background generation
in setting \bgvaesetting{}}.
%
%

\subsection{Validation on ShapeNet}
\label{sec:experiments-shapenet}

For our experiments on synthetic data, we use four ShapeNet~\cite{shapenet15arxiv} classes: car, chair, airplane, and sofa.
These have very different characteristics---cars have a smooth, largely-convex shape and areas of high-frequency texture; chairs
and airplanes have much more angular shapes with multiple elongated parts, but are often of more uniform texture;
sofas are typically concave with large, flat surfaces.
We use 80\% of the instances in each class for training, and 20\% for evaluation.

Our model is trained using renderings rather than the meshes themselves.
For car and sofa, we use the renderings of \cite{choy16eccv}; for chair and airplane, we obtained better results using those
of \cite{haene173dv}, which have greater pose diversity.
Although these datasets contain several images per mesh, we shuffle the images randomly, so there is only a very small chance of
the same object appearing twice in one minibatch. 
As a form of data augmentation, we alpha-composite the renderings over random solid color backgrounds.

\begin{table}
    \centering
    \begin{tabular}{ @{} l c c c c c c c @{} }
        \toprule
        & \multicolumn{3}{c}{\masksetting} && \multicolumn{3}{c}{\bgvaesetting} \\
        \cmidrule{2-4} \cmidrule{6-8}
        & IS & FID & KID && IS & FID & KID \\
        \midrule
        airplane & 4.0 & 73.5 & 0.063 && 3.2 & 56.5 & 0.044 \\
        car & 4.1 & 154.0 & 0.123 && 3.5 & 165.4 & 0.136 \\
        chair & 5.8 & 111.1 & 0.088 && 5.2 & 82.6 & 0.061 \\
        sofa & 4.3 & 58.3 & 0.037 && 4.3 & 63.8 & 0.041 \\
        \bottomrule
    \end{tabular}
    \vspace{2pt}
    \caption{
        Quantitative measures of generation for four ShapeNet classes, in settings \masksetting{} and \bgvaesetting{}
        (training with and without ground-truth masks respectively)
        For IS, larger is better; for FID/KID, smaller is better.
    }
    \label{tab:class-vs-setting}
\end{table}

\myparagraph{Generation with and without mask annotations.}
We train separate models on each of the four classes, in each of the two supervision settings \masksetting{} and \bgvaesetting{}.
In \fig{class-vs-setting-examples}, we see that our method has learnt to generate plausible samples for all four classes.
The samples are reasonably diverse in terms of both texture and shape---note the different colors of car, different
styles of chair, etc.
More examples are shown in \app{more-samples}.
Even in setting \bgvaesetting{}, the model has learnt to sample meshes that
represent a complete instance of the foreground object, without including any sections of background---in spite of training
without segmentation annotations.
This is particularly impressive for chairs, which have narrow legs that could easily be ignored.
Quantitatively, \tab{class-vs-setting} shows that \bgvaesetting{} does give slightly poorer results than \masksetting{} in
terms of IS, which is expected as it is a significantly more challenging setting.
For FID and KID, performance is similar in the two settings---car and sofa are better with \masksetting{}, and airplane and chair with \bgvaesetting{}.

\begin{table}
    \centering
    \begin{tabular}{ @{} l c c c c c c @{} }
        \toprule
        & \densep{} && \multicolumn{4}{c}{\nimp{}} \\
        \cmidrule{2-2} \cmidrule{4-7}
        & int. faces && int. faces & IS & FID & KID \\
        \midrule
        airplane & 56.1 \% && 0 \% & 3.7 & 76.3 & 0.067 \\
        chair & 58.8 \% && 0 \% & 5.4 & 145.8 & 0.127 \\
        sofa & 77.0 \% && 0 \% & 4.0 & 82.6 & 0.058 \\
        \bottomrule
    \end{tabular}
    \vspace{2pt}
    \caption{
        Mean fraction of intersecting faces per generated mesh (int. faces) in setting \masksetting{} using
        \densep{} and \nimp{} parametrizations.
        We also give the generation metrics for \nimp{}; see \tab{class-vs-setting} for the same using
        \densep{}.
    }
    \label{tab:pushing-vs-regular}
\end{table}

\begin{figure}
    \centering
    \includegraphics[height=2cm]{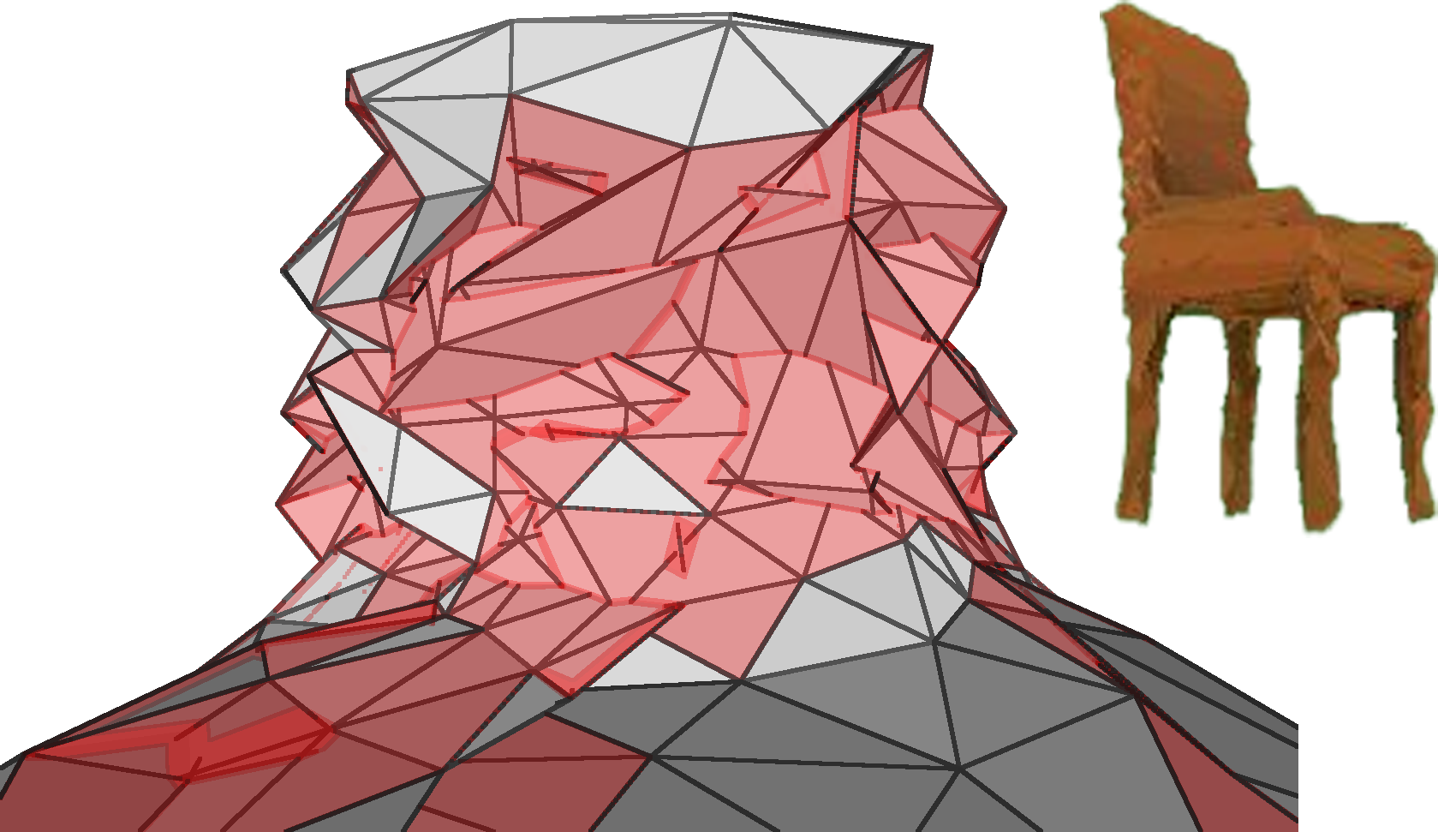}\hspace{0.5cm}
    \includegraphics[height=2cm]{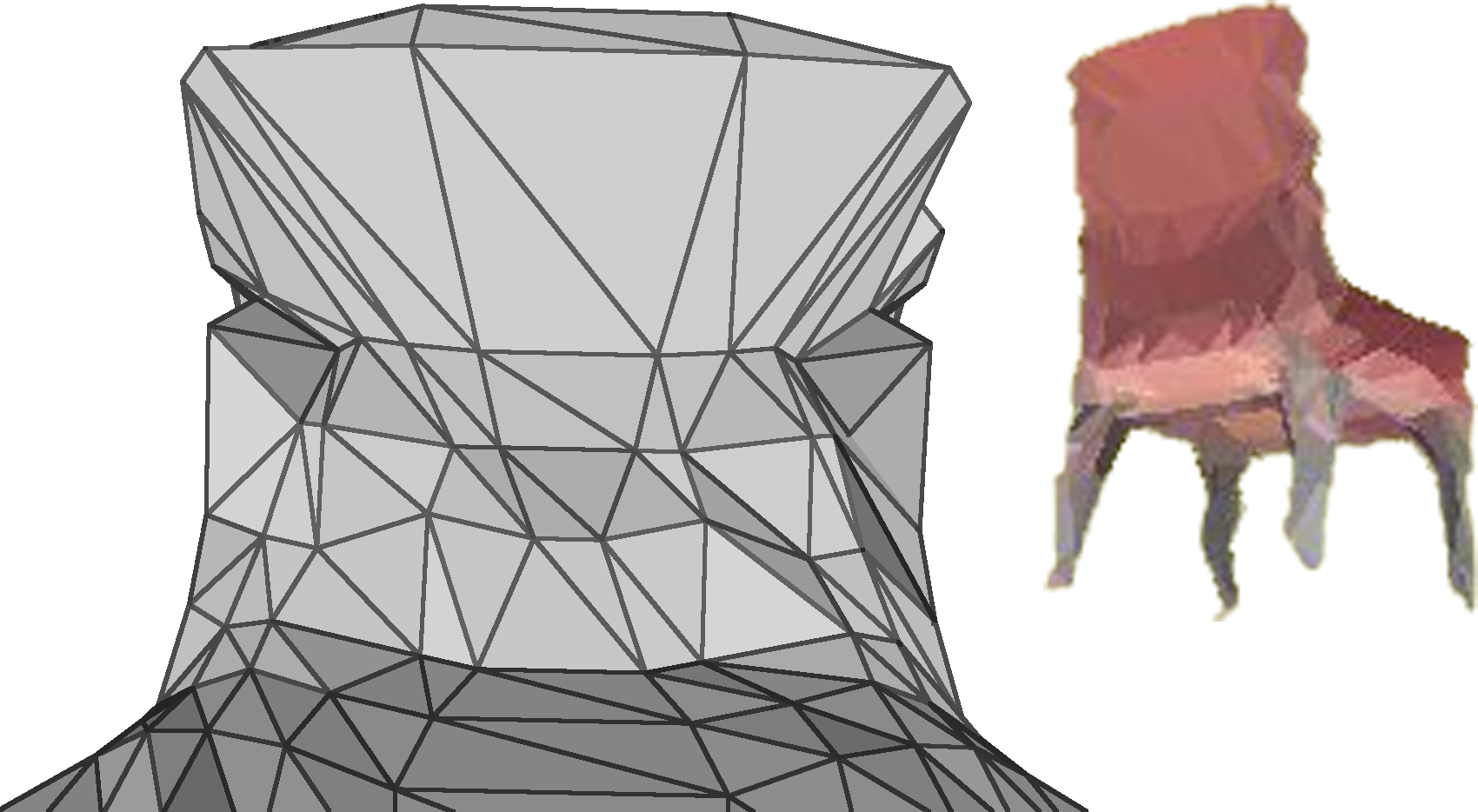}
    \caption{
        Examples of chair mesh structure and renderings using densely-connected \densep{} (left) and non-intersecting
        \nimp{} (right) mesh parametrizations.
        Faces that intersect others are highlighted in red.
        Note that the right-hand mesh is free of intersecting faces, in contrast to the left. Consistency
        of the renderings with an input image is not sufficient to enforce this, as both renders appear reasonable.
        In both cases, we used the same strengths for the local smoothness regularizers, so these do not influence the result.
    }
    \label{fig:pushing-vs-regular-examples}
\end{figure}

\begin{figure*}
    \centering
    \begin{tabular}{c@{\hspace{1.5cm}}c}
        \masksetting{} & \bgvaesetting{} \\
        \\
        \includegraphics[width=0.45\linewidth]{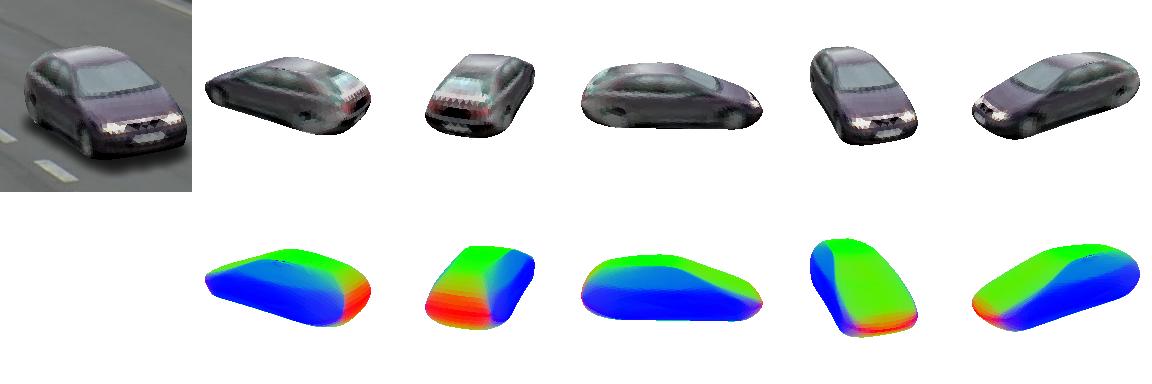} &
        \includegraphics[width=0.45\linewidth]{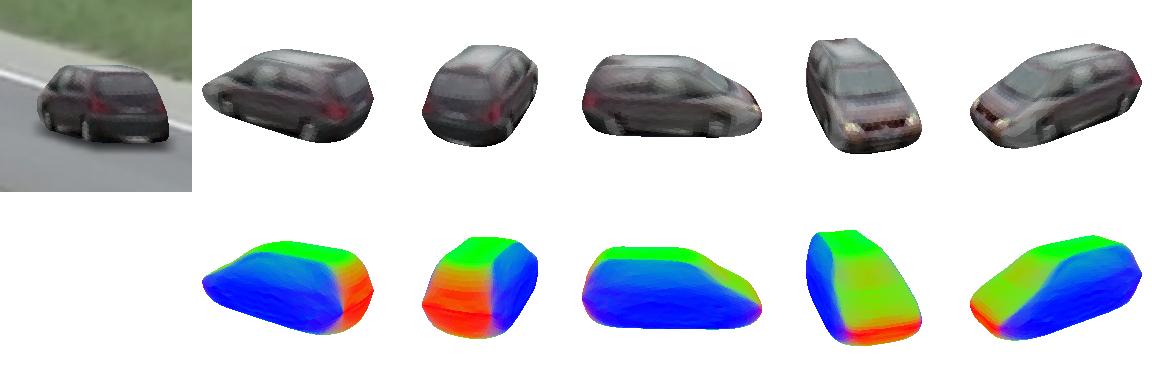}\\
        \includegraphics[width=0.45\linewidth]{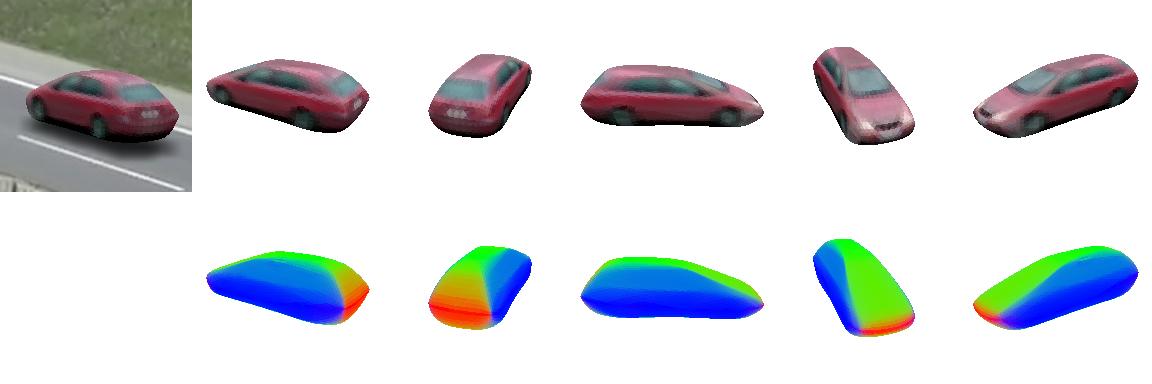} &
        \includegraphics[width=0.45\linewidth]{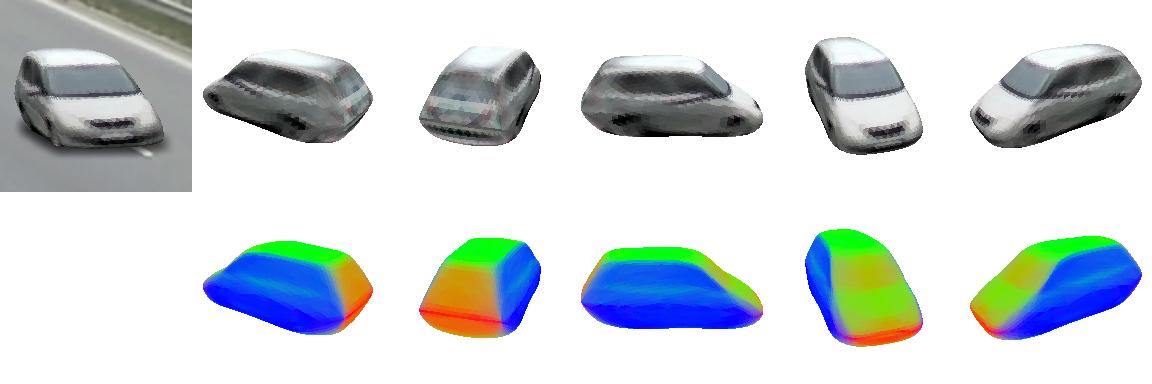}\\
        \includegraphics[width=0.45\linewidth]{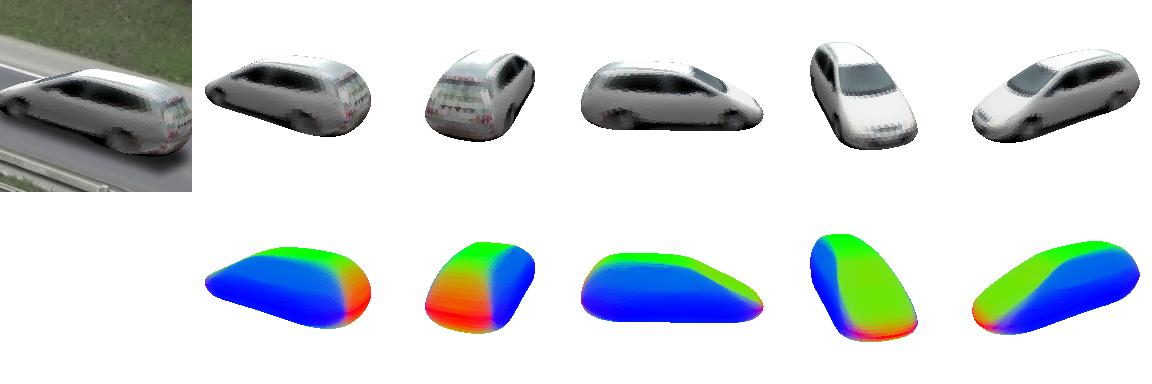} &
        \includegraphics[width=0.45\linewidth]{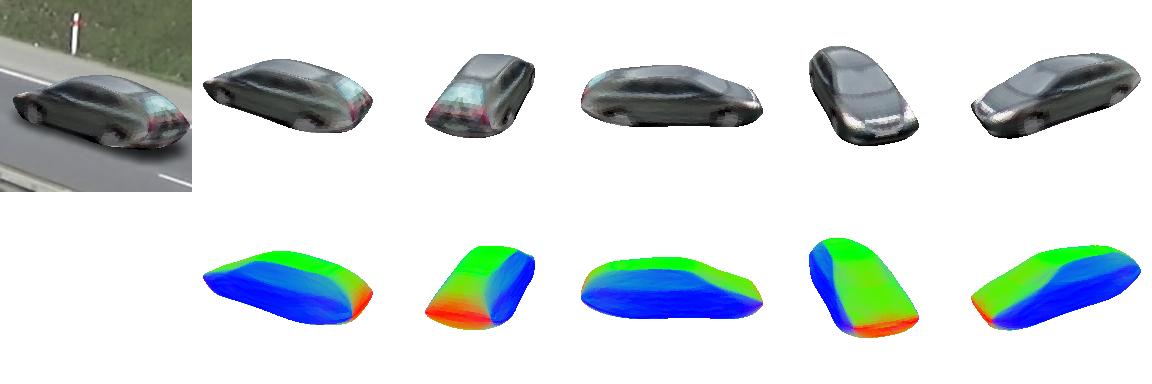}\\
        \includegraphics[width=0.45\linewidth]{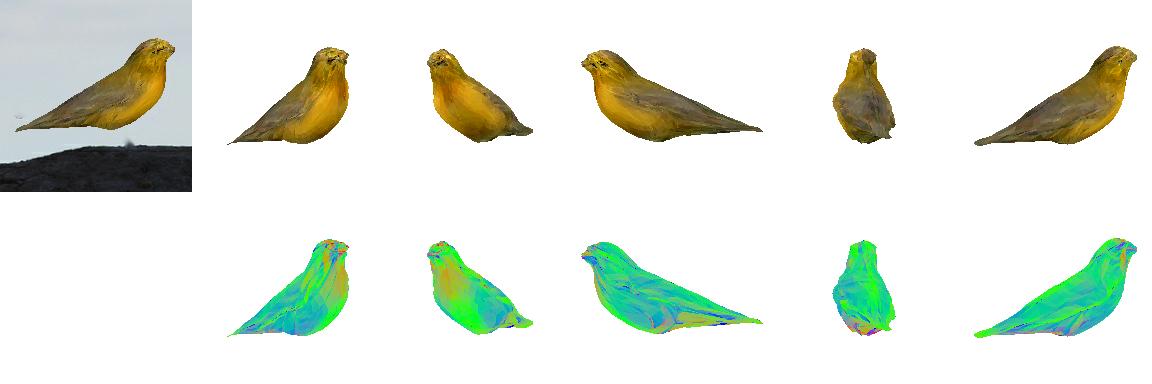} &
        \includegraphics[width=0.45\linewidth]{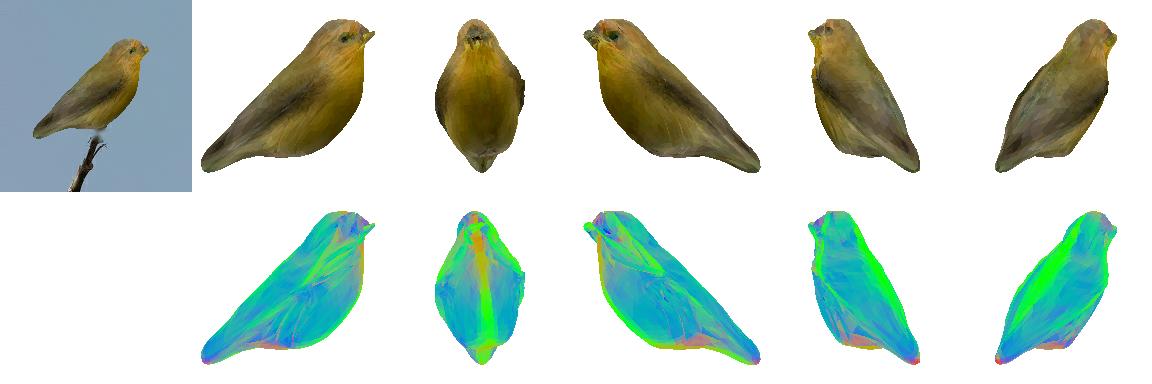}\\
        \includegraphics[width=0.45\linewidth]{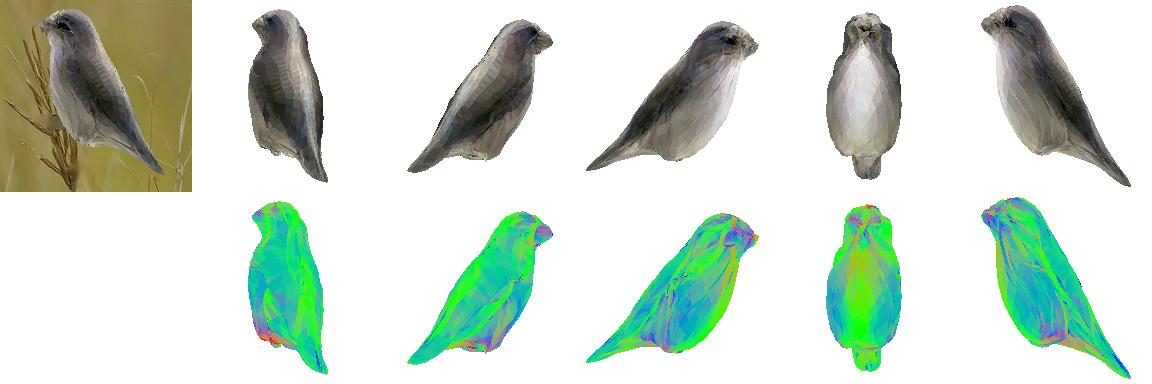} &
        \includegraphics[width=0.45\linewidth]{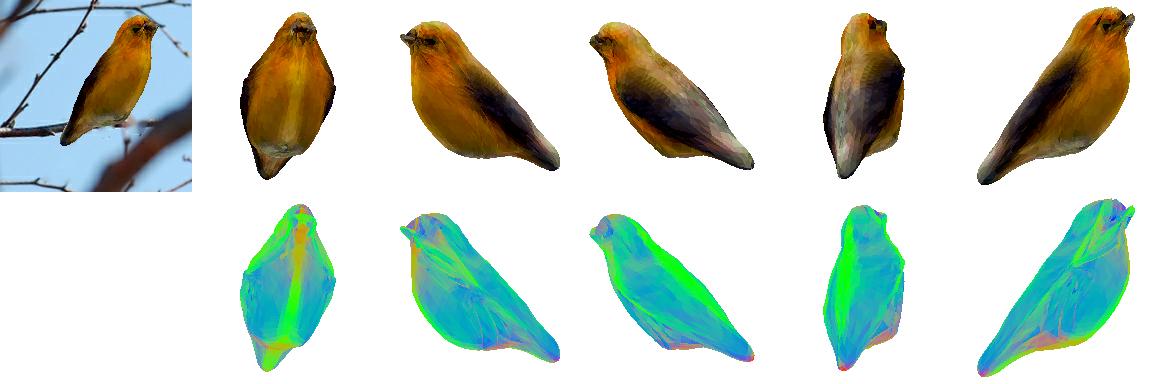}\\
        \includegraphics[width=0.45\linewidth]{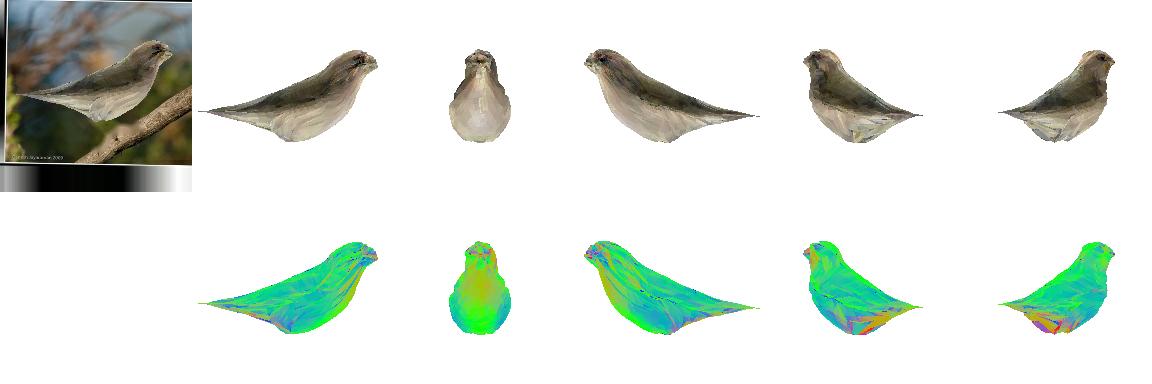} &
        \includegraphics[width=0.45\linewidth]{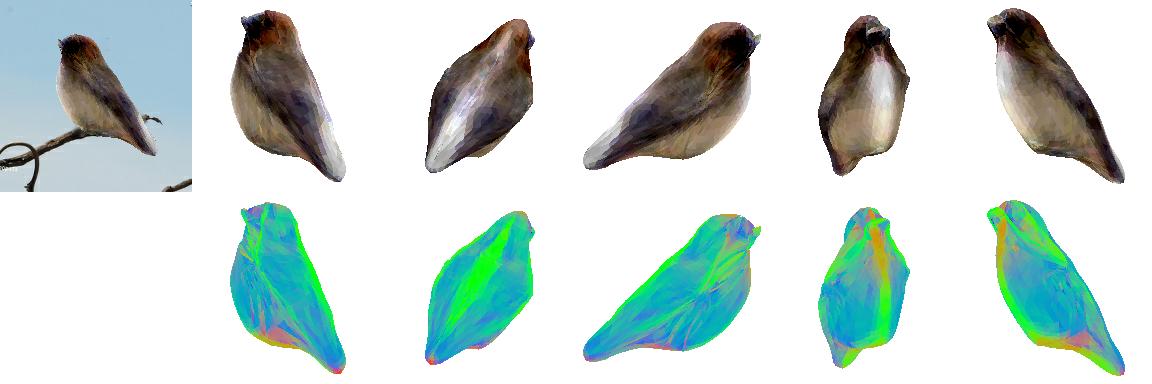}
    \end{tabular}
    \caption{
    Cars and birds generated by our model.
    Each group of 11 images shows different views of one textured 3D mesh sampled from our model.
    The left-hand image shows the sampled mesh rendered with a background and pose drawn randomly
    from the test dataset.
    The lower five images in each group are normal maps, which reveal the 3D structure more clearly.
    Samples in the the left column use setting \masksetting{}; those in the right column use \bgvaesetting{}.
    }
    \label{fig:car-and-bird-examples}
\end{figure*}

\begin{figure*}
    \centering
    \begin{tabular}{c@{\hspace{1cm}}c}
        \masksetting{} & \bgvaesetting{} \\
        \\
        \includegraphics[width=0.47\textwidth]{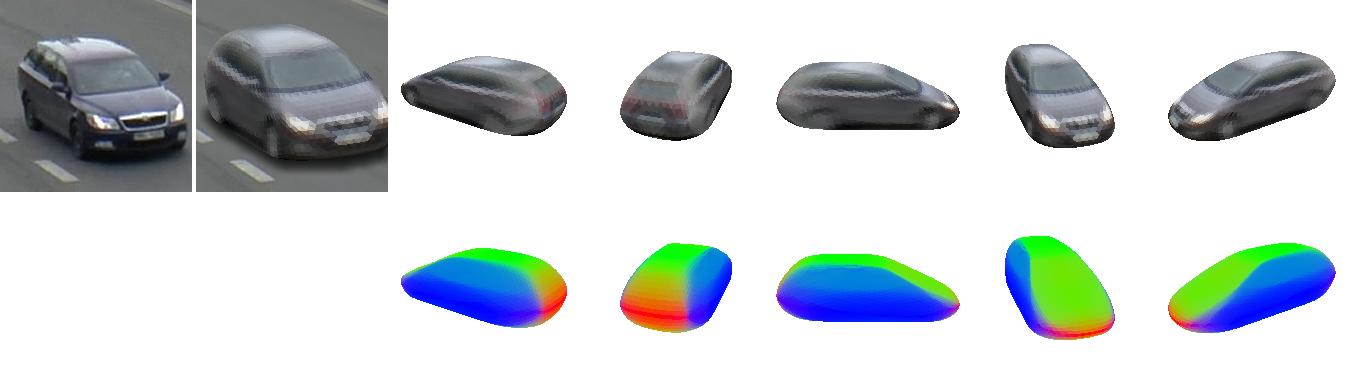} &
        \includegraphics[width=0.47\textwidth]{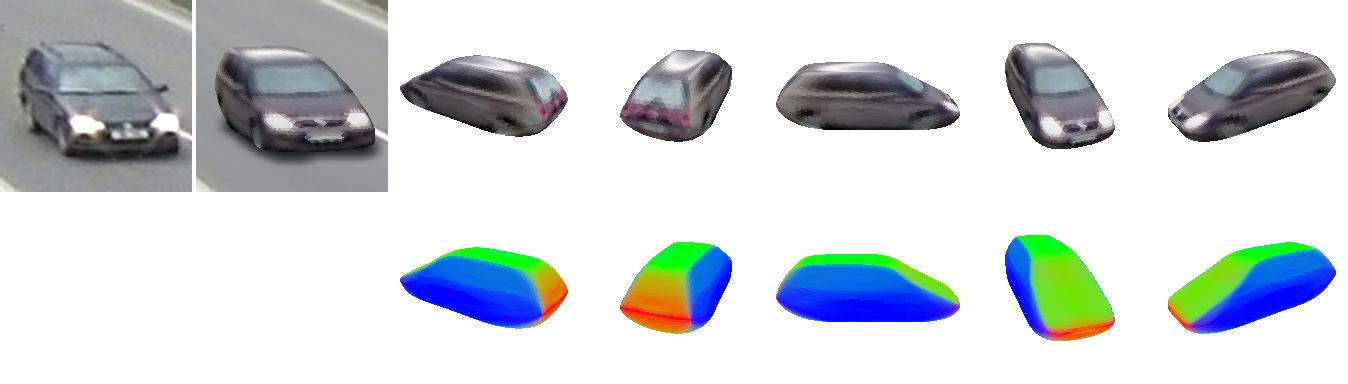} \\
        \includegraphics[width=0.47\textwidth]{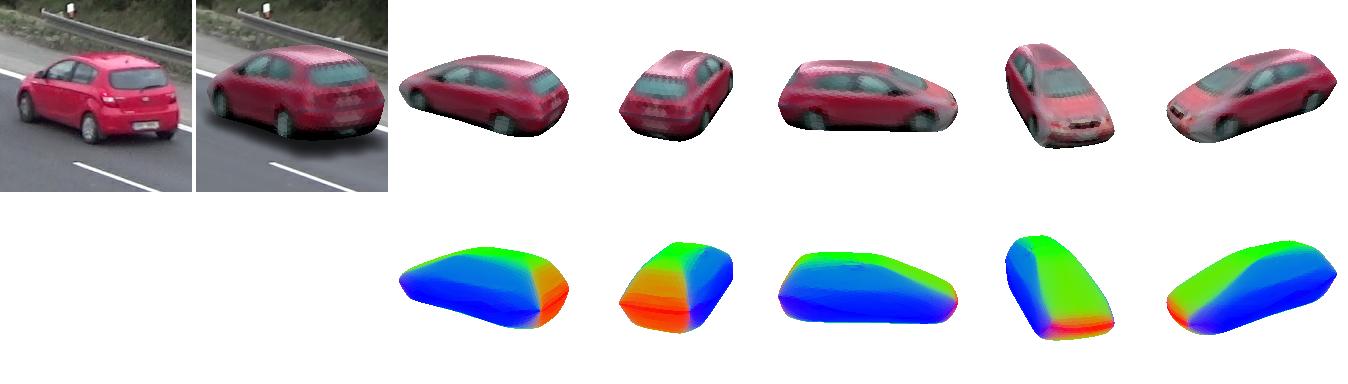} &
        \includegraphics[width=0.47\textwidth]{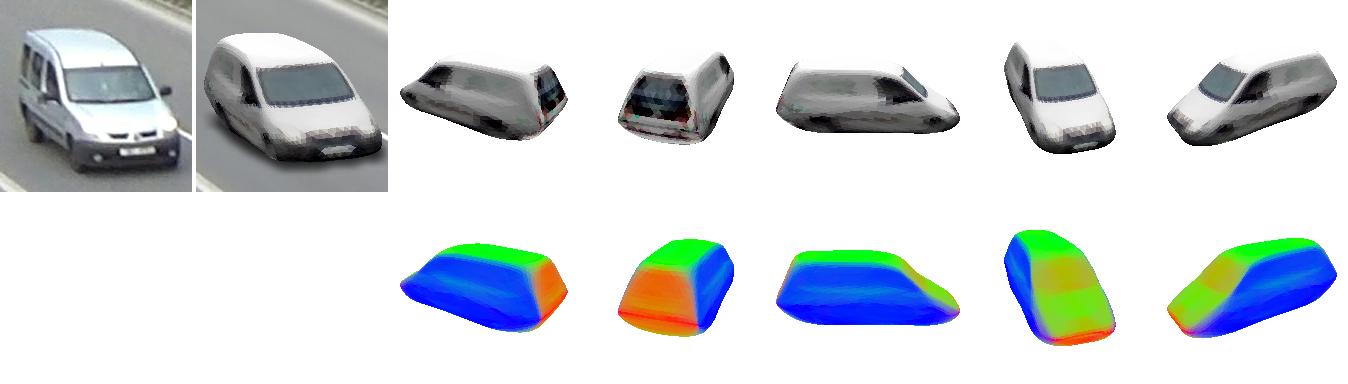} \\
        \includegraphics[width=0.47\textwidth]{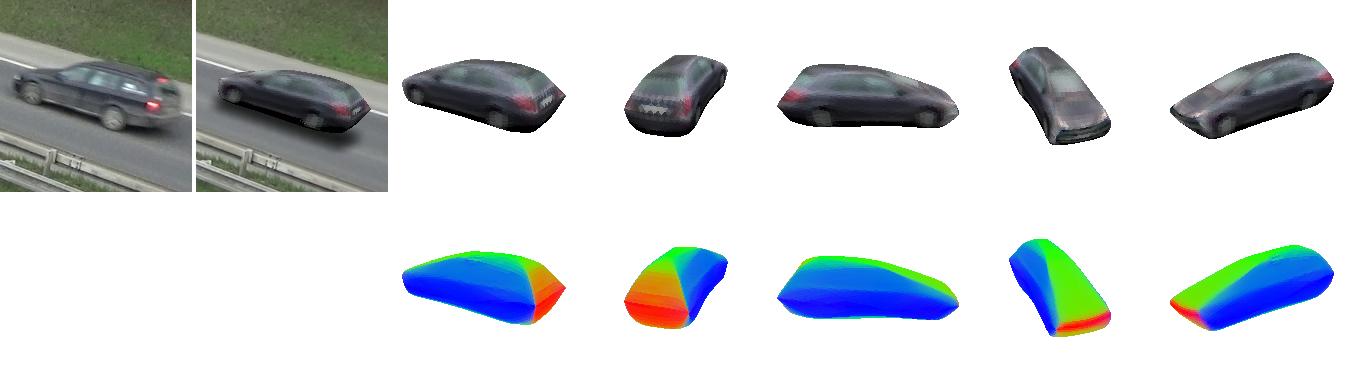} &
        \includegraphics[width=0.47\textwidth]{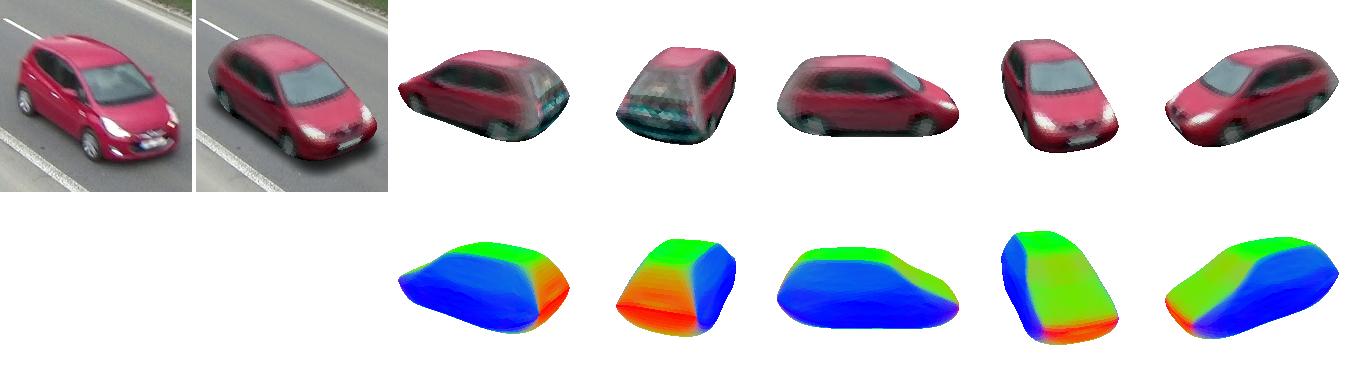} \\
        \includegraphics[width=0.47\textwidth]{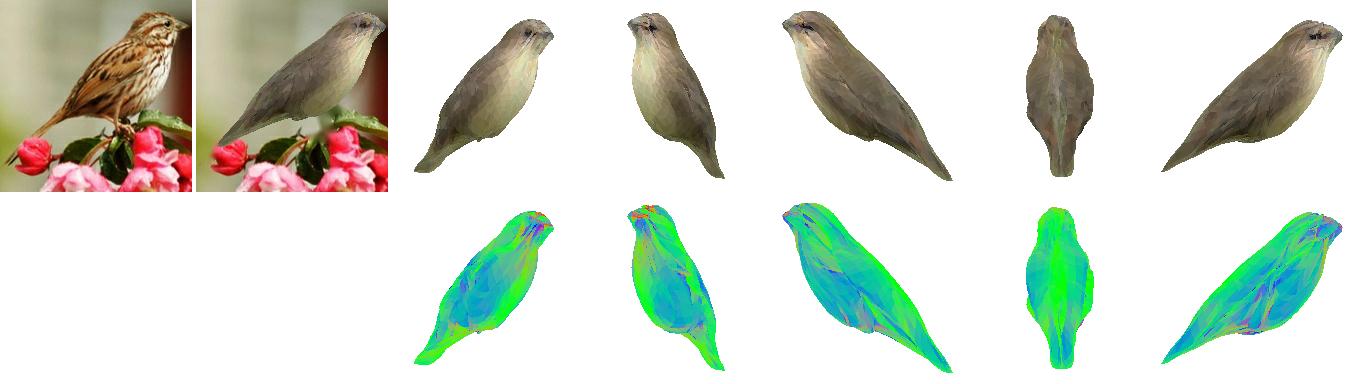} &
        \includegraphics[width=0.47\textwidth]{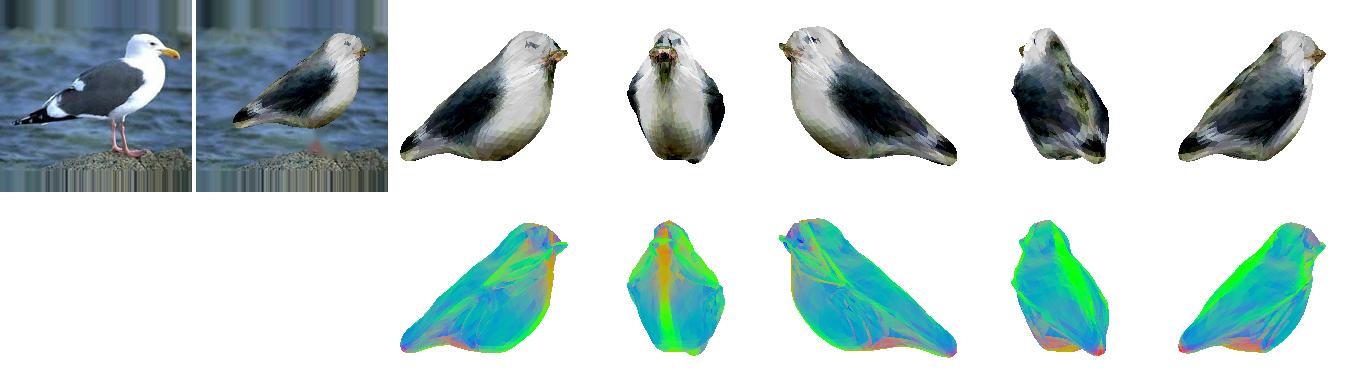} \\
        \includegraphics[width=0.47\textwidth]{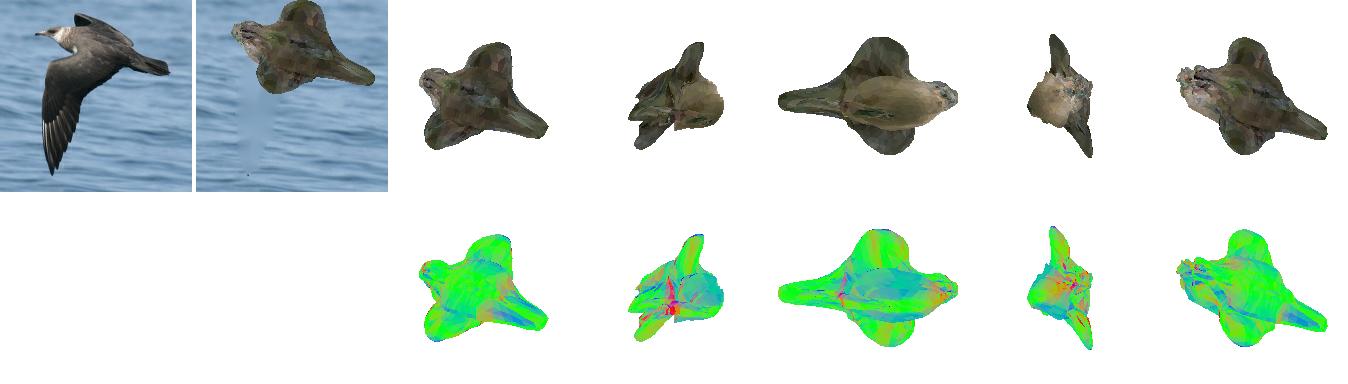} &
        \includegraphics[width=0.47\textwidth]{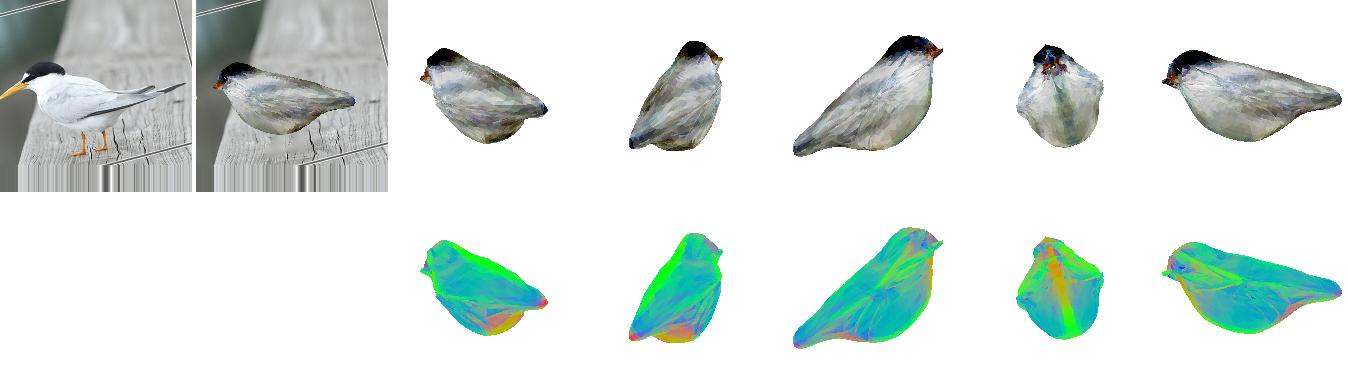} \\
        \includegraphics[width=0.47\textwidth]{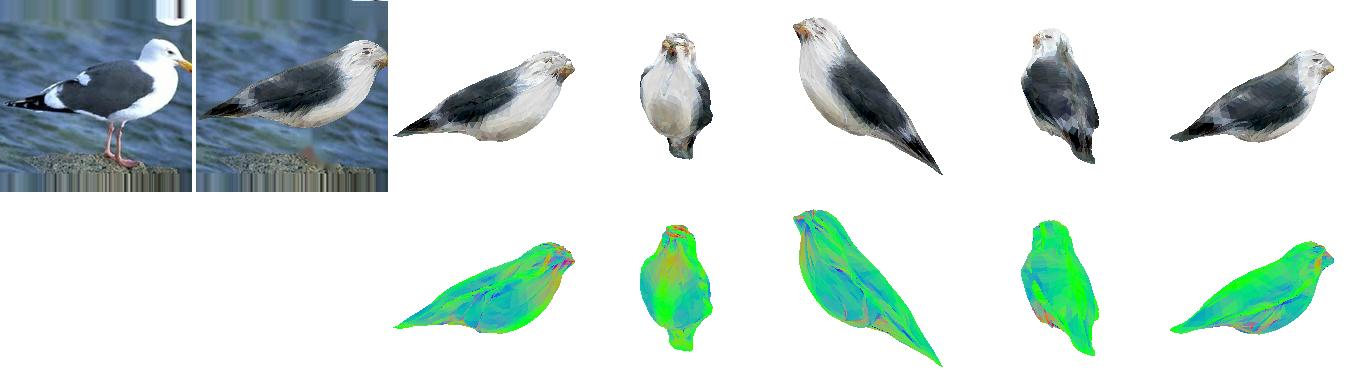} &
        \includegraphics[width=0.47\textwidth]{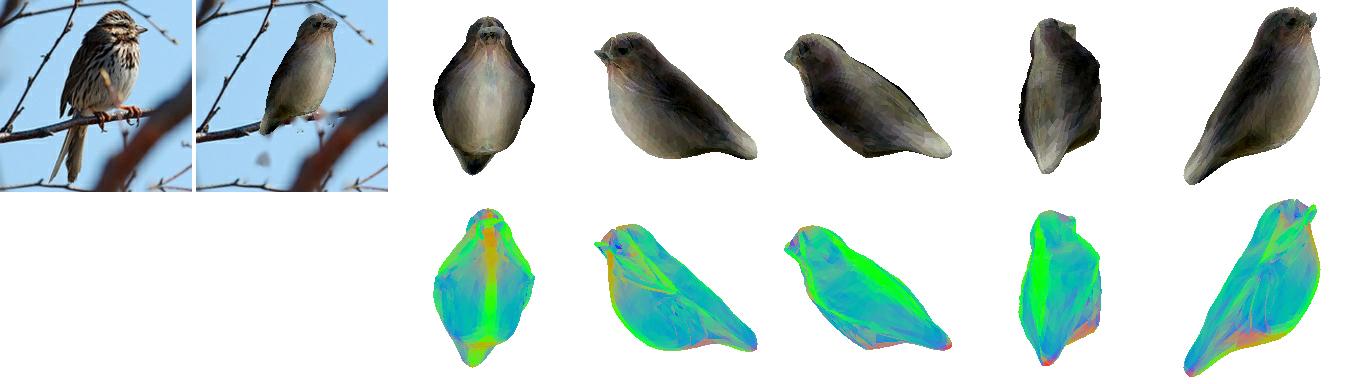}
    \end{tabular}
    \caption{
    Textured 3D reconstructions of cars and birds.
    The left-hand image in each group is the input to our model; the next is our reconstruction, rendered over the
    ground-truth background; the remaining five columns are different views of the reconstructed mesh, with
    normal-maps below.
    Samples in the the left column use setting \masksetting{}; those in the right column use \bgvaesetting{}.
    }
    \label{fig:car-and-bird-recon}
\end{figure*}

\myparagraph{Non-intersecting mesh parametrization}
We now evaluate our non-intersecting mesh parametrization \nimp{}, comparing it to a simple dense decoder \densep{}
that directly outputs vertex locations.
Qualitative examples with \nimp{} are given in \fig{class-vs-setting-examples} for classes chair and airplane; we see
that this parametrization is powerful enough to produce realistic shapes, in spite of it having considerably
fewer degrees of freedom than \densep{}.
\fig{pushing-vs-regular-examples} shows details of two chairs for illustration, one sampled with \densep{} and the other
with \nimp{}.
While both render to give reasonable images, the sample with \densep{} has numerous intersecting faces visible, while that
with \nimp{} does not.

For quantitative comparison, we use the classes chair, airplane, and sofa, for which local surface regularizers struggle to prevent
self-intersections.
As well as the three generation metrics, we also measure the mean fraction of faces that intersect another face. 
%
We see (\tab{pushing-vs-regular}) that \densep{} typically produces shapes with substantial numbers of self-intersections, while
\nimp{} avoids these.
Thus, meshes from \nimp{} may be used in downstream tasks which require a non-intersecting surface, in contrast to those
from \densep{}.
%
Conversely, all three generation metrics indicate slightly worse performance with \nimp{}.
We believe this 
is because
although \nimp{} is highly expressive, the decoder network is less free to make arbitrary movements to individual
vertices in response to gradients, as vertices are tied together through the shared directions \direction{}.
The space of shapes is therefore more difficult to traverse by gradient descent.
%
Thus, there is a trade-off to be made depending on the desired application---when non-intersection is important, \nimp{} should be
used, but when it is not a requirement, \densep{} may be more appropriate.



\begin{table}
    \centering
    \begin{tabular}{ @{} l c c c @{~~}c c c c @{} }
        \toprule
        & \multicolumn{3}{c}{\masksetting} && \multicolumn{3}{c}{\bgvaesetting} \\
        \cmidrule{2-4} \cmidrule{6-8}
        & IS & FID & KID && IS & FID & KID \\
        \midrule
        \textit{Our model} \\
        ~~~car & 2.8 & 191.4 & 0.211 && 3.0 & 182.1 & 0.197 \\
        ~~~bird & 4.5 & 104.3 & 0.090 && 3.8 & 75.4 & 0.060 \\[4pt]
        \textit{2D VAE} \\
        ~~~car & - & - & - && 3.0 & 157.4 & 15.4 \\
        ~~~bird &  - & - & -  && 3.3 & 213.8 & 21.0 \\
        \bottomrule
    \end{tabular}
    \vspace{0pt}
    \caption{
        Quantitative measures of generation, training on natural images of cars and birds, in settings \masksetting{} and
        \bgvaesetting{}.
        For IS, larger is better; for FID/KID, smaller is better.
        For reference, we also include results from a 2D deep convolutional VAE.
    }
    \label{tab:cars-and-birds}
\end{table}

\subsection{Generation results on natural images}
\label{sec:experiments-real}


For our experiments on natural images, we use two classes: bird and car.
The images of birds are taken from the CUB-200-2011 dataset~\cite{WahCUB_200_2011}; we use the preprocessed version
from \cite{kanazawa18eccv}, including their approximate camera calibrations, and discard the class labels.
Each image contains one bird, which we crop out using the provided segmentation masks, applying a random offset for data augmentation.
For setting \masksetting{}, we synthesise a background by inpainting the region corresponding to the bird,
replacing each pixel inside the mask with the nearest non-foreground pixel.
The images of cars are taken from the BrnoCompSpeed dataset~\cite{sochor18tits}, and we use their approximate camera
calibration.
We segment and crop out individual cars using \cite{girshick18detectron}, and use the median frame as the background.
Although the original data is video, we sample frames sparsely and shuffle them, treating them as independent images.
%

We see (\fig{car-and-bird-examples}) that our approach yields realistic 3D meshes for both classes
(more samples are shown in \app{more-samples}).
%
In setting \masksetting{}, the sampled cars have diverse shapes and colors; in the more challenging setting \bgvaesetting{},
the shapes are a little less diverse, and the colours a little more blurry, but all samples are still clearly identifiable as cars.
For birds, the results are even stronger; there is no perceptible decrease in performance when we do not have mask annotations.
This is perhaps because the highly varied backgrounds of the bird images are difficult for the foreground model to (incorrectly)
incorporate.

A similar pattern is apparent in the quantitative results (\tab{cars-and-birds}), where we see that \bgvaesetting{} birds in fact perform
significantly better than any other combination. 
Meanwhile, for car, we see that \bgvaesetting{} performs slightly worse than \masksetting{}, in accordance with the qualitative results.
We also give results for a conventional (2D) deep convolutional VAE, with similar latent capacity, encoder architecture, and training time as
our models, to give a reference for the ranges of the different metrics.
Our 3D models give significantly better results than this baseline for birds, and somewhat worse for cars; this possibly reflects
the high degree of background complexity in CUB.

\subsection{Single-image 3D reconstruction on natural images}
\label{sec:experiments-recon}

While trained for generation, our model learns single-image 3D reconstruction `for free'---we can obtain a textured mesh reconstruction by
running the encoder and decoder networks (green and blue parts of \fig{model}) on an image.
Specifically, the encoder network yields a variational posterior distribution on the latent variables;
we take the mode of this distribution, and pass it through
the decoders $\mathrm{dec}_\mathrm{shape}$ and $\mathrm{dec}_\mathrm{color}$ to produce a textured 3D mesh.
Thus, we perform single-image 3D reconstruction, in spite of our model never having received any
3D supervision, nor even segmentation masks in \bgvaesetting{} setting.
Note also that we did \textit{not} tune our models for reconstruction quality.

\fig{car-and-bird-recon} shows examples for each class in both training settings,
on held-out validation images; more examples are given in \app{more-samples}.
For the bird images, our setting \masksetting{} matches that of \cite{kanazawa18eccv}, allowing qualitative comparison
with their approach~\footnote{the background we use to illustrate the reconstruction for birds is generated
from the original image, with all points inside the (ground-truth) bird mask replaced by the nearest non-masked pixel}.
%

We see that in almost all cases, the network faithfully reconstructs the visible part of the object, including
details of texture and shape.
Notably, occluded regions are also reconstructed plausibly, even though they do not influence the photometric reconstruction loss.
This is because the model must learn to produce textured shapes that reproject well for all training images,
while representing their variability in a low-dimensional latent space.
Moreover, it must do so subject to the KL term in the loss (\ref{eq:loss}) that explicitly limits the latent capacity.
This discourages solutions where the predicted shapes explain each image in isolation,
but occluded parts have appearances lying outside the space of variability that is observed in images where that part is visible.



\section{Conclusion}
\label{sec:discussion}

We have presented a new generative model of textured 3D meshes, and shown how to train this from images alone,
by augmenting it to capture the entire image formation process.
We train the model to explain its training images, by reconstructing each in terms of a foreground mesh rendered
over a background.
We have shown that this approach allows us to generate diverse and realistic textured meshes of five object classes.
Importantly, our method can use natural images, not just renderings, as training data.
Moreover, it does not rely on multiple views of the same instance, nor on ground-truth segmentation masks.
The resulting meshes are textured, and so may be used immediately in downstream applications such as visual
effects and games.


We have also presented a new mesh parametrization, that avoids intersections \textit{a priori}.
This is useful whenever we need to generate a mesh from a neural decoder with the guarantee
that it does not contain any self-intersections, necessary for example if it is to be used for physical simulation
or 3D printing.
However, this comes at the expense of producing samples that score slightly lower than a na\"{i}ve parametrization
in terms of IS/FID/KID metrics.

%
%
%
%

{\small
\bibliographystyle{ieee_fullname}
\bibliography{shortstrings,calvin,vggroup}
}

\appendix

\section{Additional qualitative results}
\label{app:more-samples}

In this section, we present additional qualitative results from our models using different parametrizations
and training settings, on each of the five object classes.

Random (uncurated) samples generated by the models trained on ShapeNet renderings of
cars are given in \fig{sh-car-gen-mask} and \fig{sh-car-gen-bgvae};
chairs in \fig{sh-chair-gen-mask-dense}, \fig{sh-chair-gen-mask-nimp} and \fig{sh-chair-gen-bgvae};
airplanes in \fig{sh-airplane-gen-mask-dense}, \fig{sh-airplane-gen-mask-nimp} and \fig{sh-airplane-gen-bgvae}; and
sofas in \fig{sh-sofa-gen-mask-dense}, \fig{sh-sofa-gen-mask-nimp} and \fig{sh-sofa-gen-bgvae}.
See \sect{experiments-shapenet} for discussion of these results.

\begin{figure*}
    \centerfloat
    \includegraphics[width=0.55\textwidth]{bcs_mask_1e2-L2_1e2-crease_32-ced/0000_no-shadow.jpg}
    \includegraphics[width=0.55\textwidth]{bcs_mask_1e2-L2_1e2-crease_32-ced/0001_no-shadow.jpg} \\
    \includegraphics[width=0.55\textwidth]{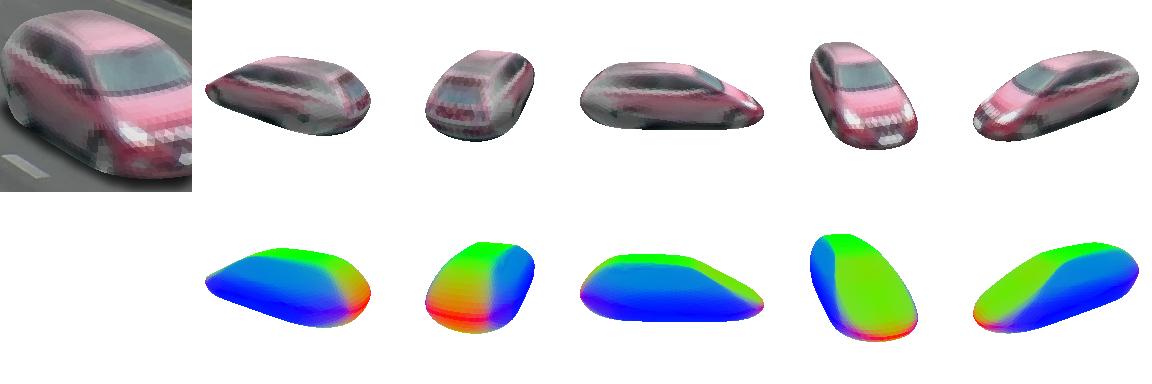}
    \includegraphics[width=0.55\textwidth]{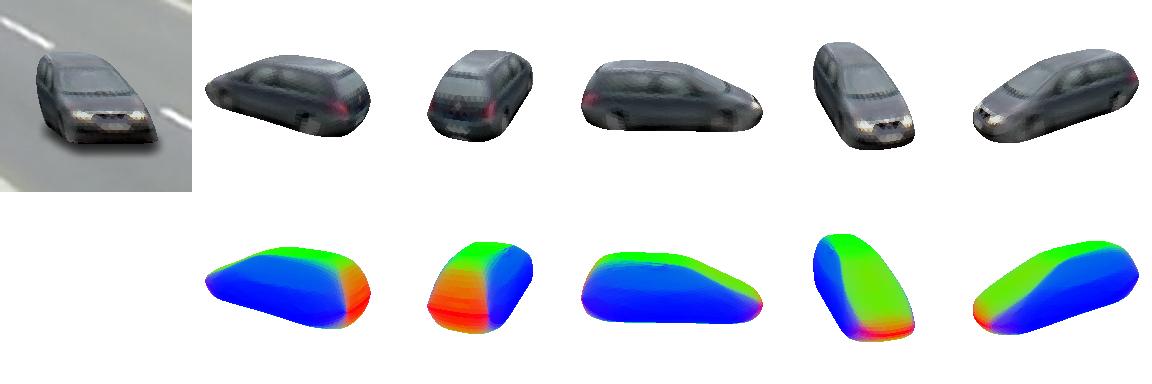} \\
    \includegraphics[width=0.55\textwidth]{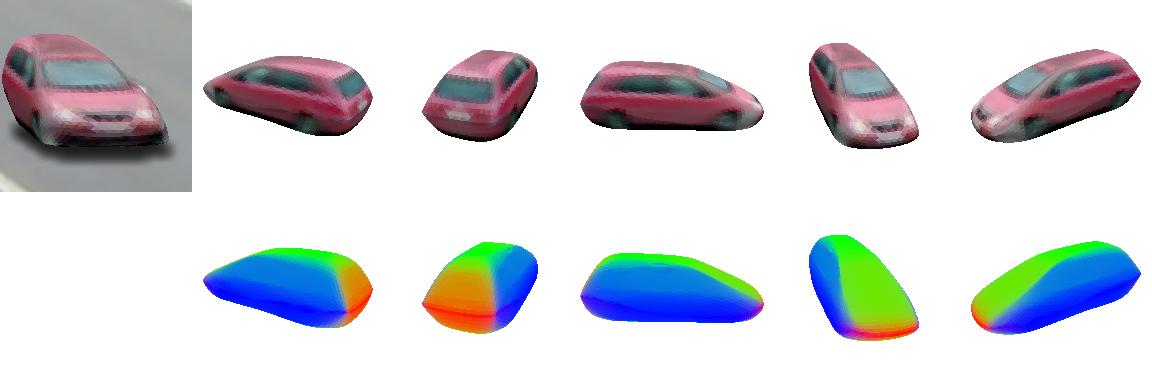}
    \includegraphics[width=0.55\textwidth]{bcs_mask_1e2-L2_1e2-crease_32-ced/0005_no-shadow.jpg} \\
    \includegraphics[width=0.55\textwidth]{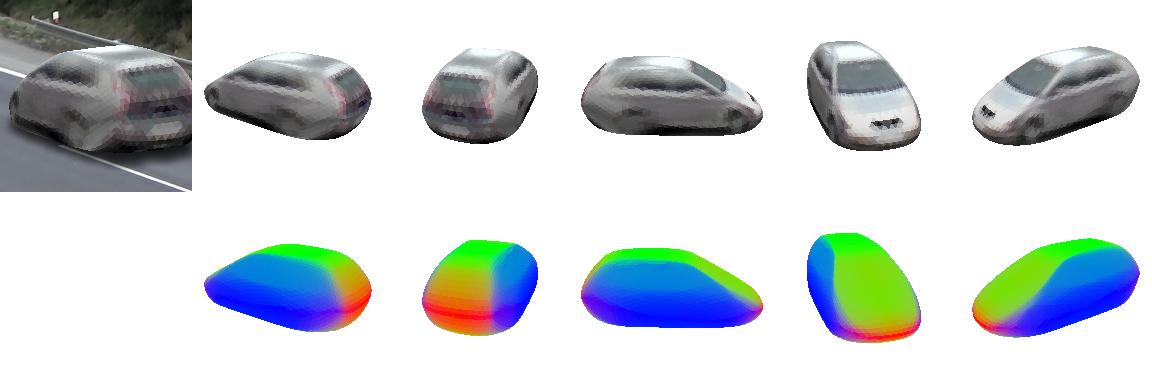}
    \includegraphics[width=0.55\textwidth]{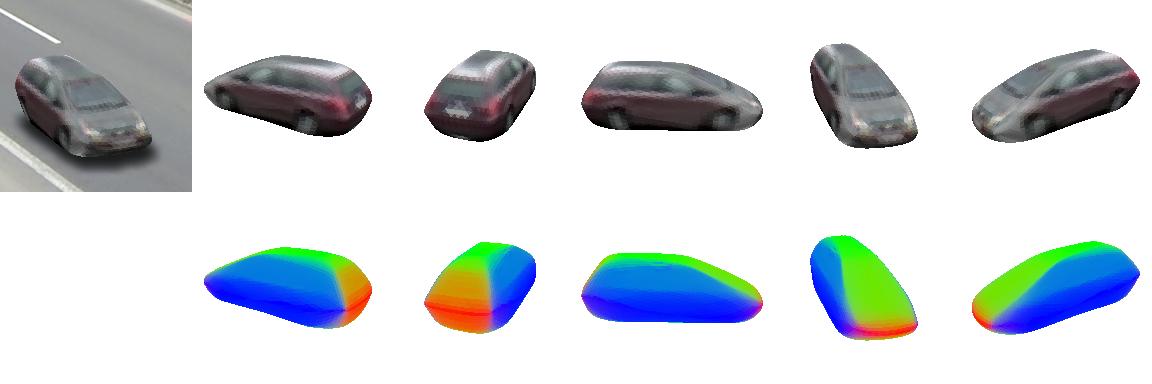} \\
    \includegraphics[width=0.55\textwidth]{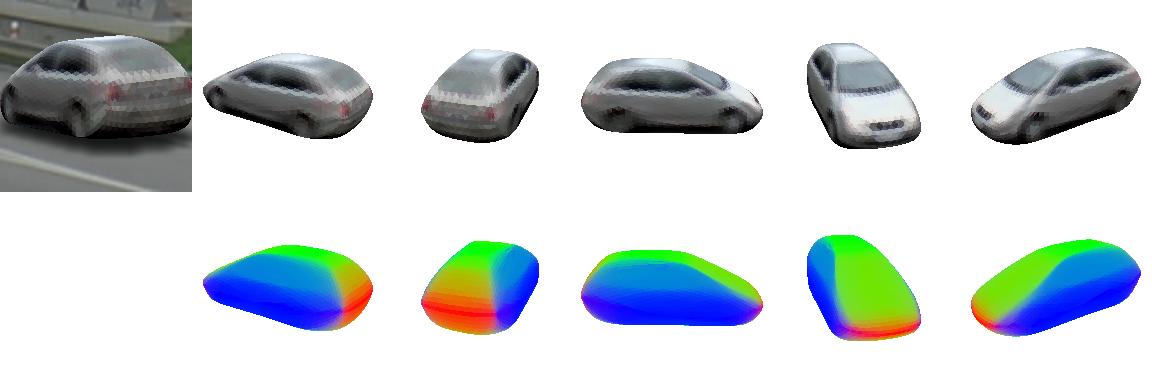}
    \includegraphics[width=0.55\textwidth]{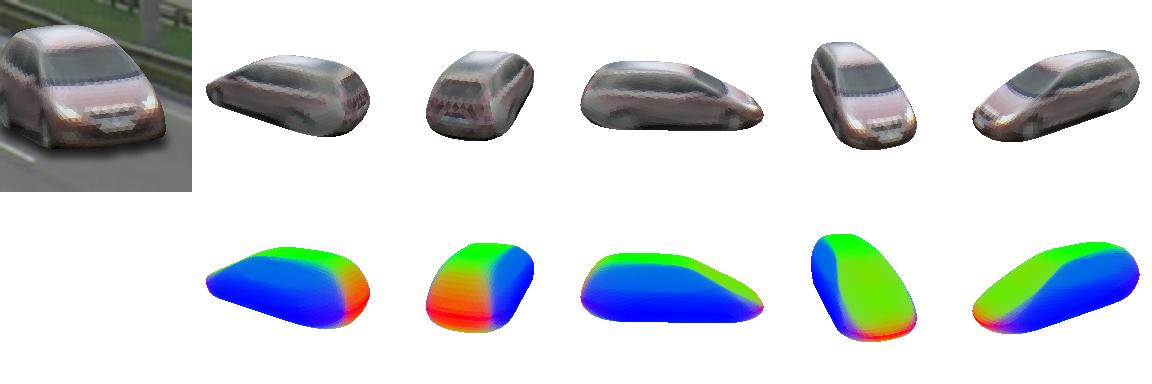} \\
    \includegraphics[width=0.55\textwidth]{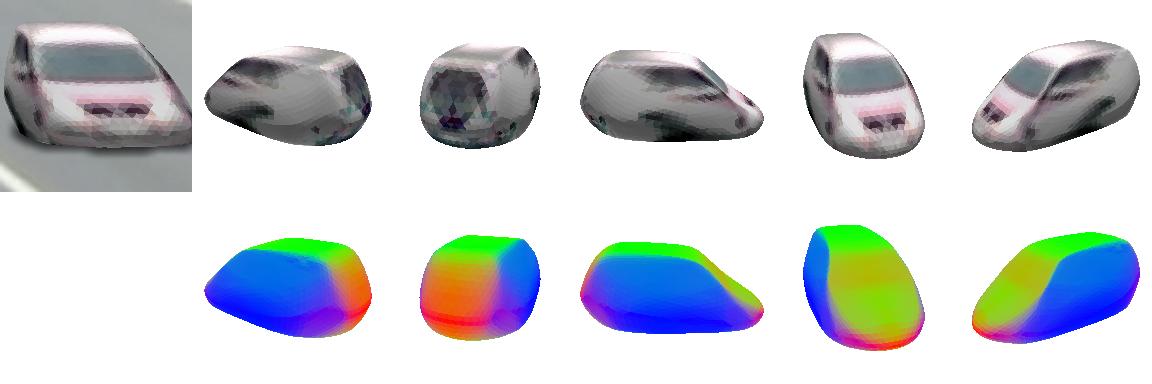}
    \includegraphics[width=0.55\textwidth]{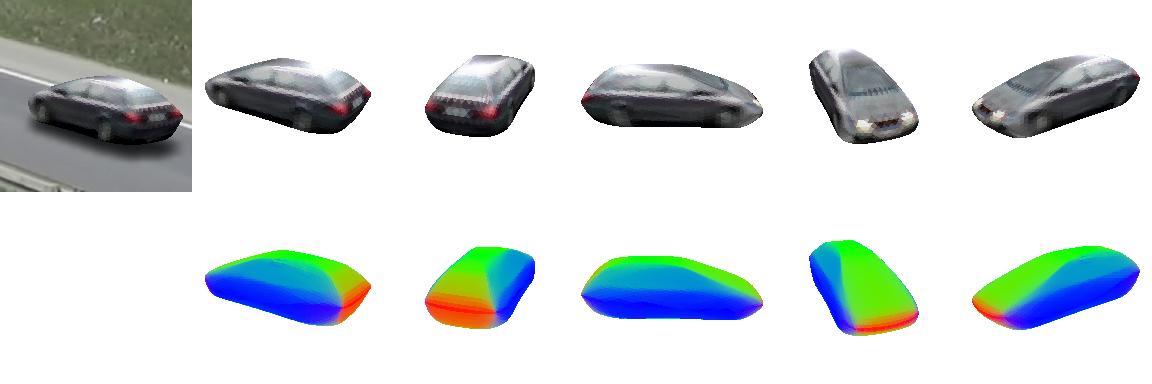}
    \caption{Examples of cars generated by our model, in setting \masksetting{}.}
    \label{fig:car-gen-mask}
\end{figure*}

\begin{figure*}
    \centerfloat
    \includegraphics[width=0.55\textwidth]{bcs_bg-vae_5e1-L2_1e1-crease/0010_no-shadow.jpg}
    \includegraphics[width=0.55\textwidth]{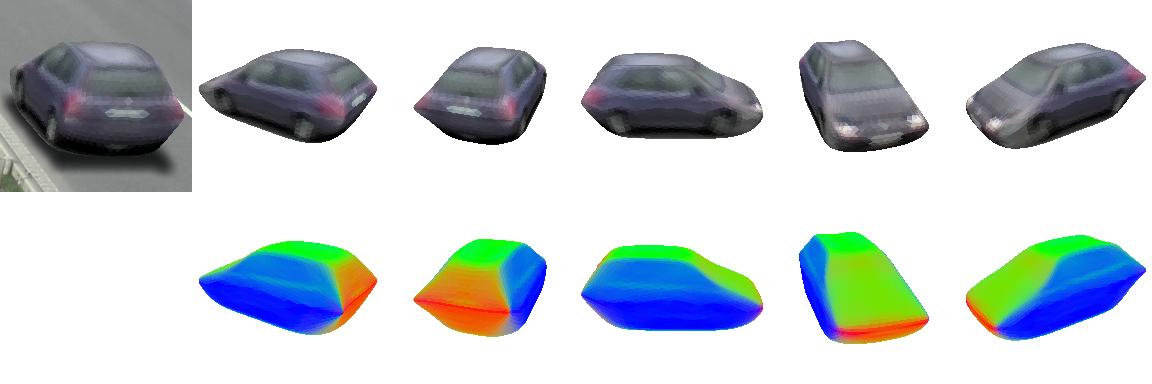} \\
    \includegraphics[width=0.55\textwidth]{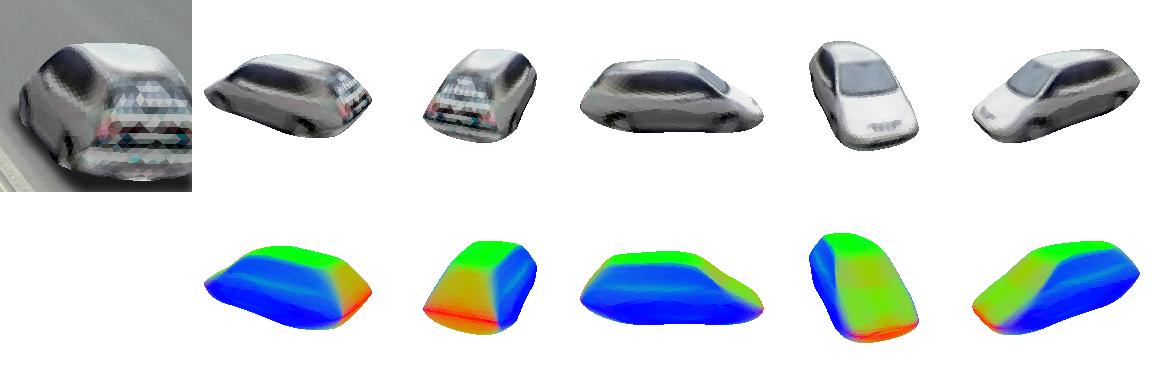}
    \includegraphics[width=0.55\textwidth]{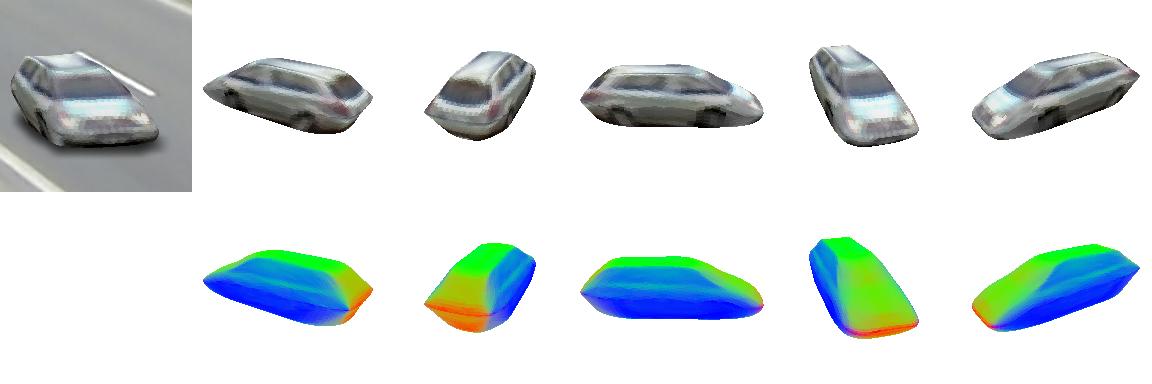} \\
    \includegraphics[width=0.55\textwidth]{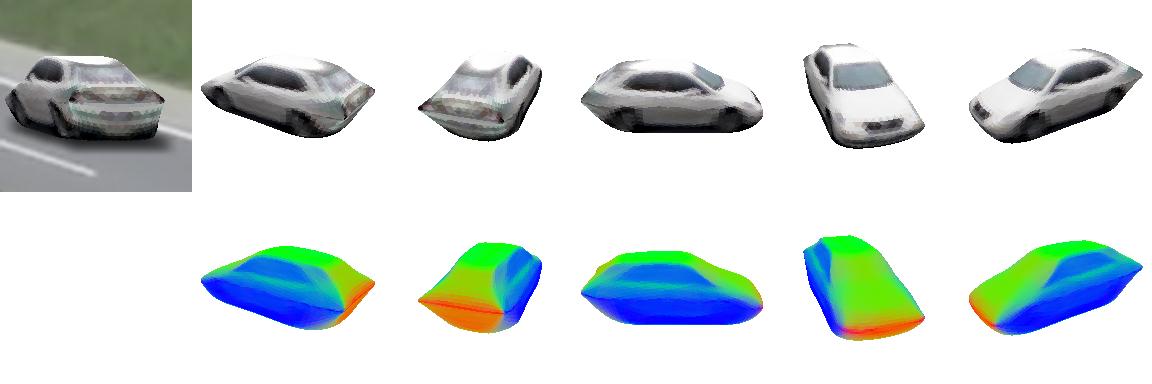}
    \includegraphics[width=0.55\textwidth]{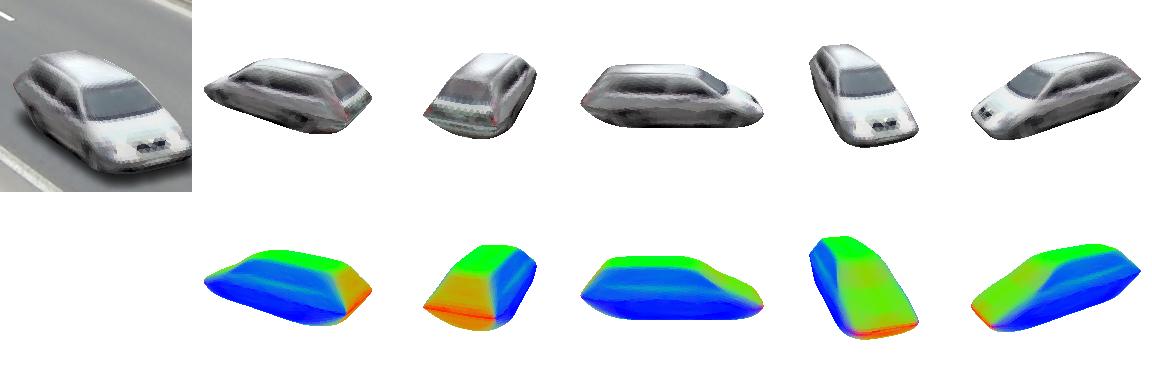} \\
    \includegraphics[width=0.55\textwidth]{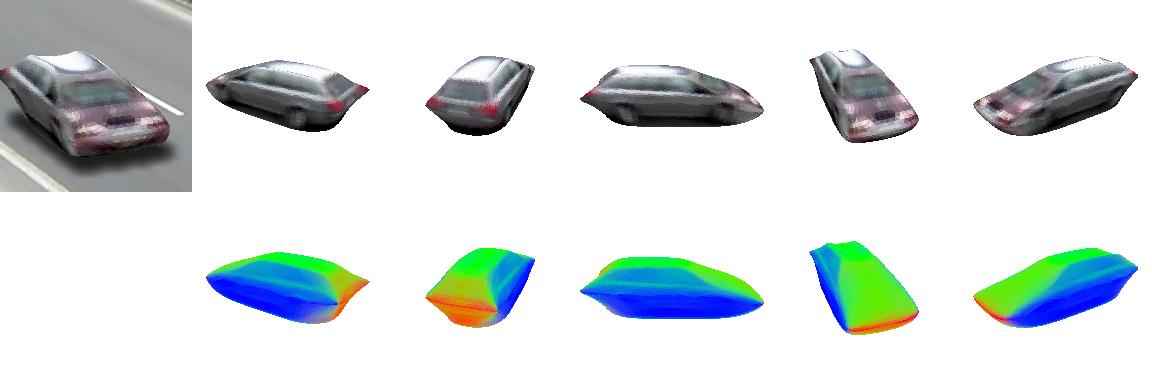}
    \includegraphics[width=0.55\textwidth]{bcs_bg-vae_5e1-L2_1e1-crease/0017_no-shadow.jpg} \\
    \includegraphics[width=0.55\textwidth]{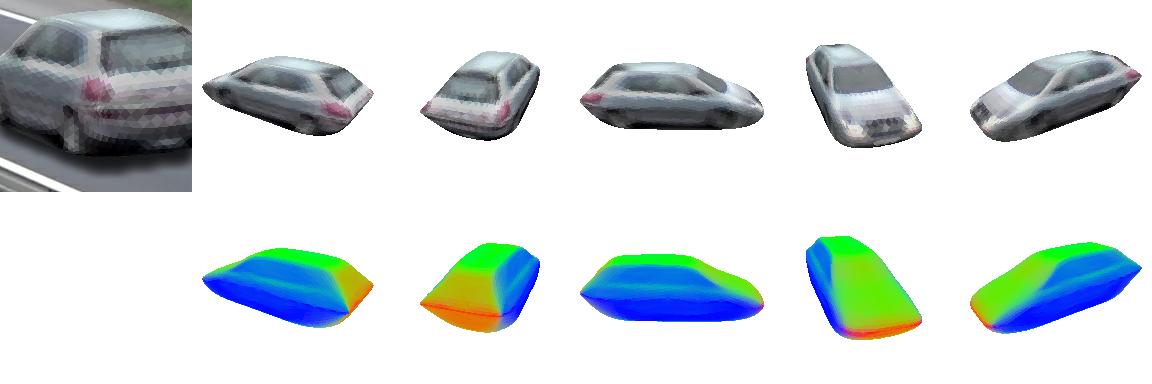}
    \includegraphics[width=0.55\textwidth]{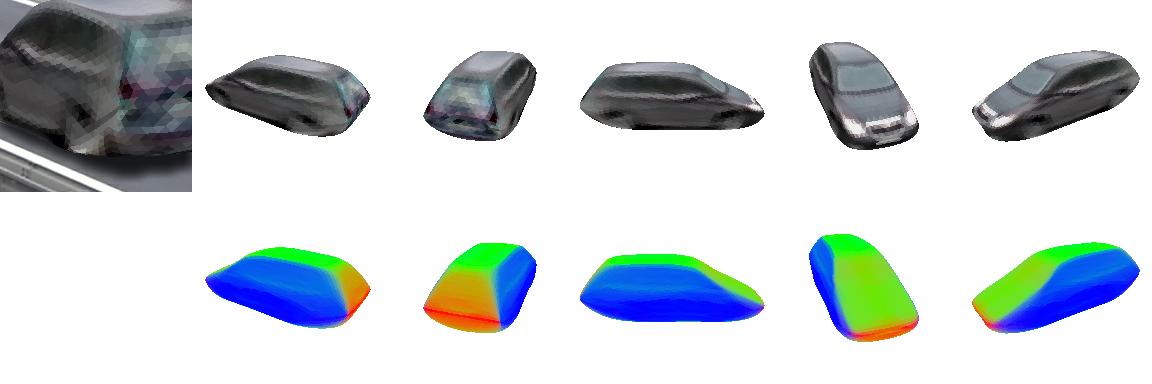} \\
    \includegraphics[width=0.55\textwidth]{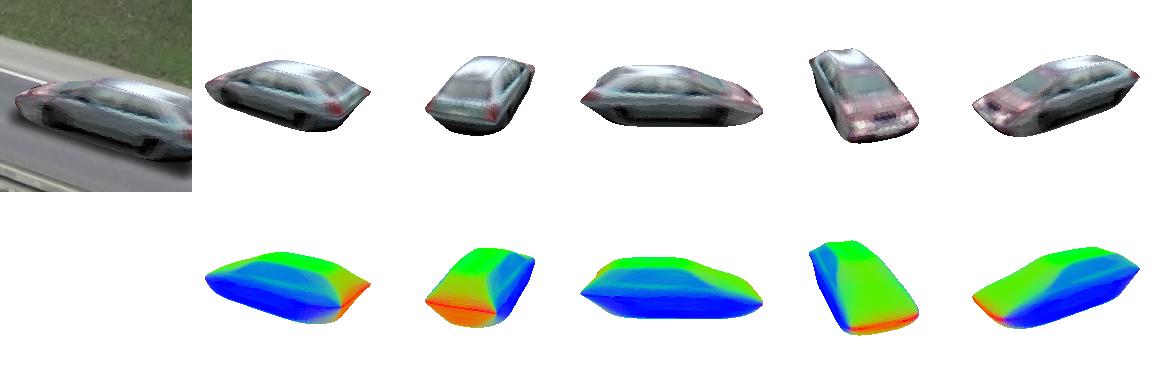}
    \includegraphics[width=0.55\textwidth]{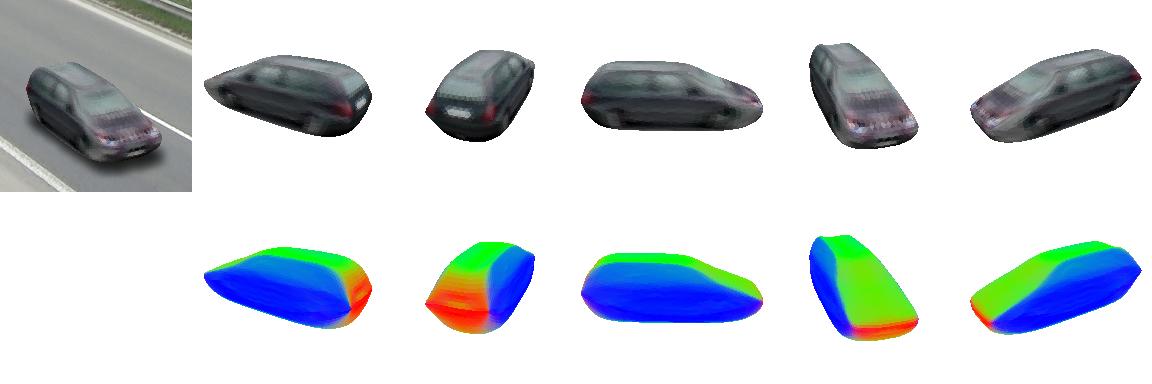}
    \caption{Examples of cars generated by our model, in setting \bgvaesetting{}.}
    \label{fig:car-gen-bgvae}
\end{figure*}

\begin{figure*}
    \centerfloat
    \includegraphics[width=0.55\textwidth]{cub_mask_1e1-L2_1e0-crease/0020_no-shadow.jpg}
    \includegraphics[width=0.55\textwidth]{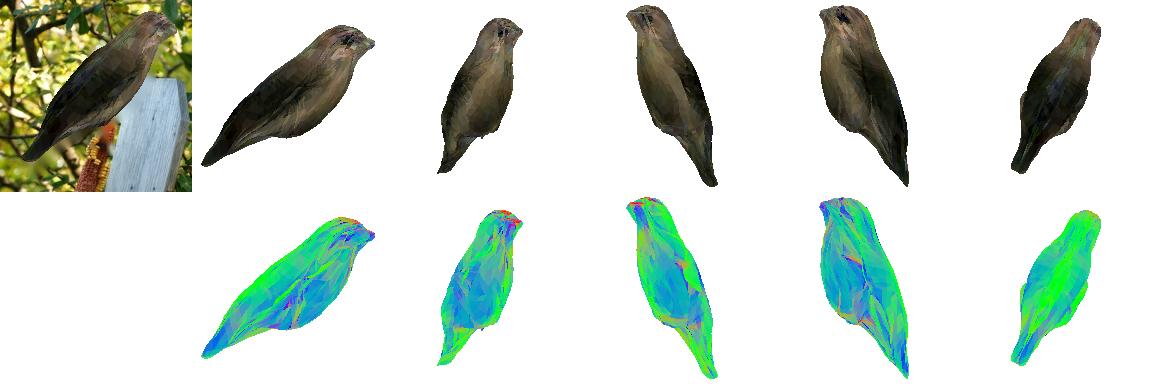} \\
    \includegraphics[width=0.55\textwidth]{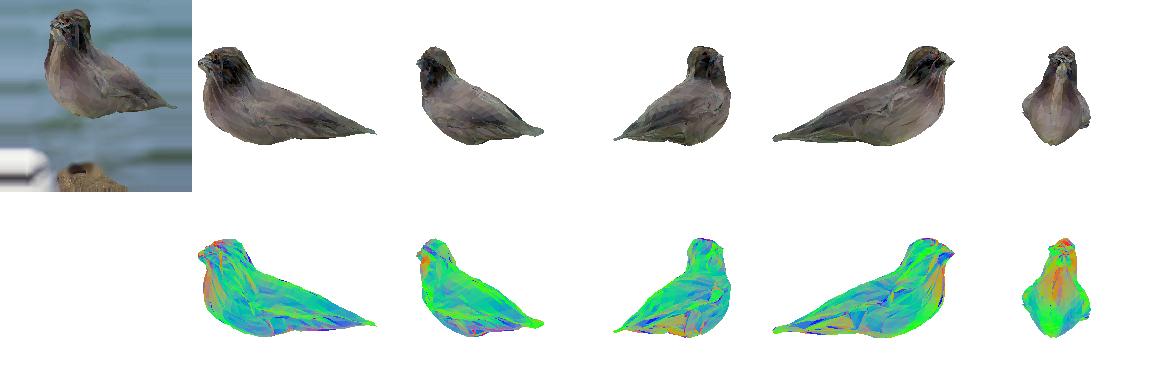}
    \includegraphics[width=0.55\textwidth]{cub_mask_1e1-L2_1e0-crease/0023_no-shadow.jpg} \\
    \includegraphics[width=0.55\textwidth]{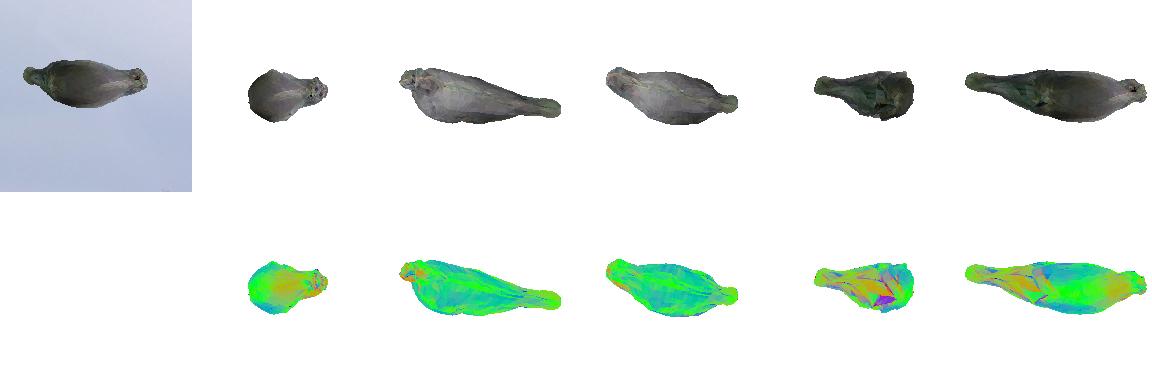}
    \includegraphics[width=0.55\textwidth]{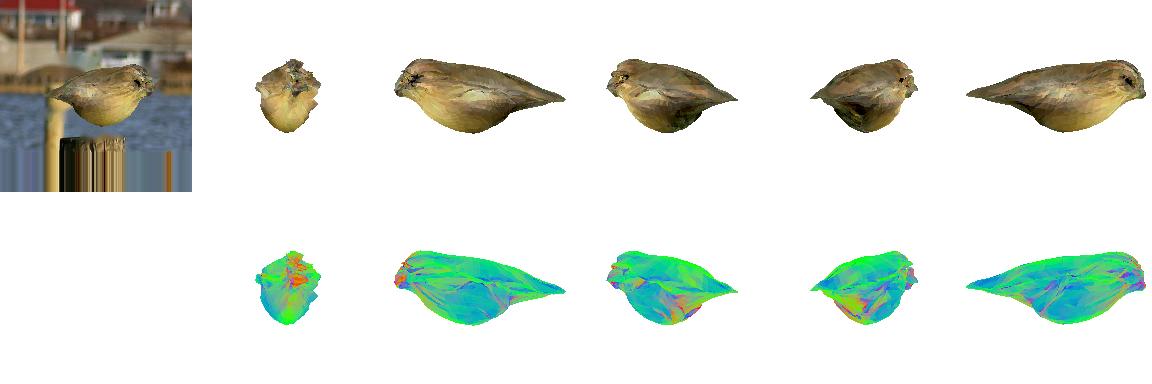} \\
    \includegraphics[width=0.55\textwidth]{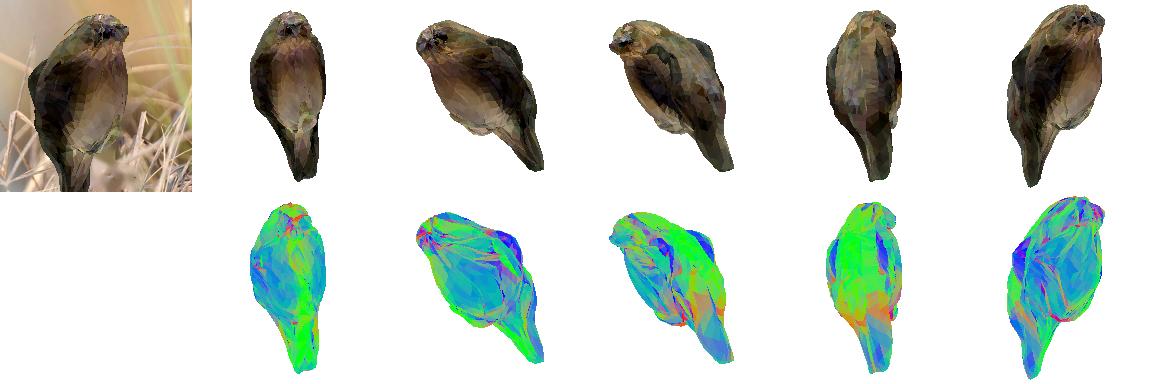}
    \includegraphics[width=0.55\textwidth]{cub_mask_1e1-L2_1e0-crease/0027_no-shadow.jpg} \\
    \includegraphics[width=0.55\textwidth]{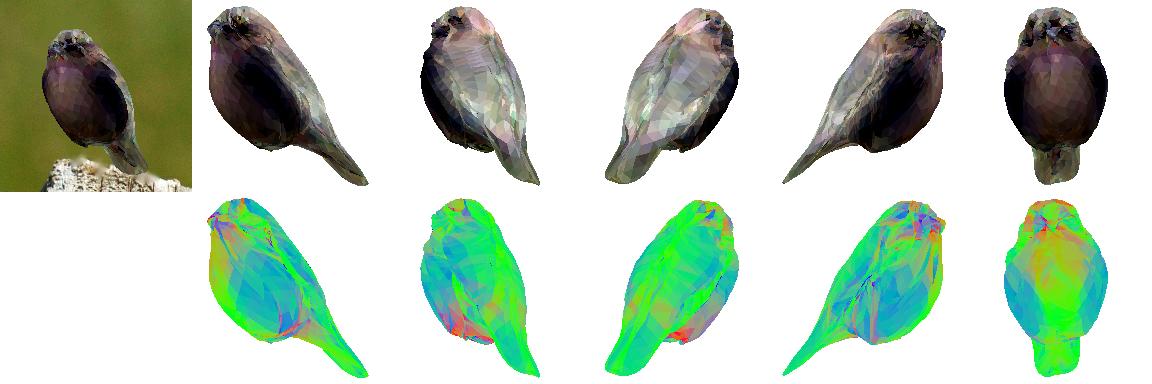}
    \includegraphics[width=0.55\textwidth]{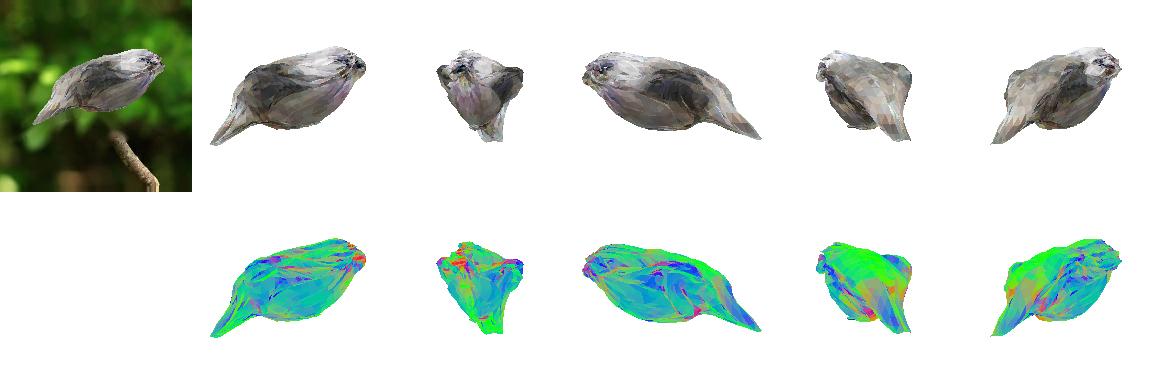} \\
    \includegraphics[width=0.55\textwidth]{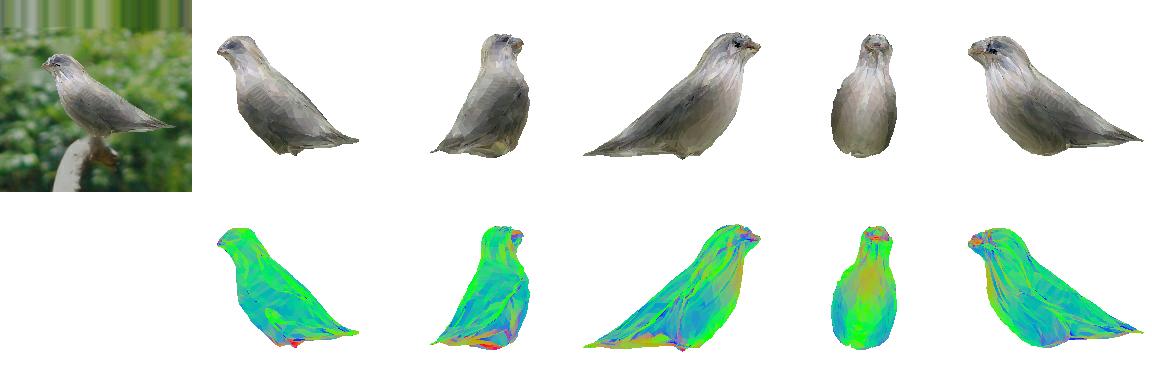}
    \includegraphics[width=0.55\textwidth]{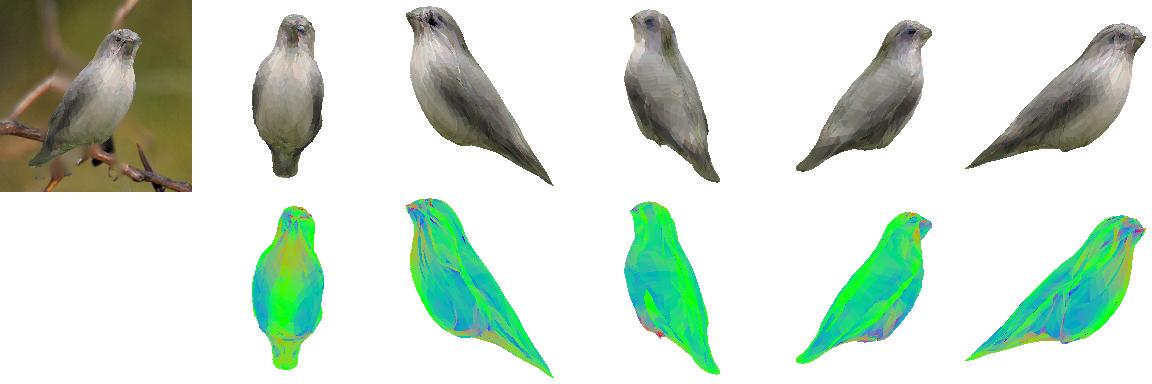}
    \caption{Examples of birds generated by our model, in setting \masksetting{}.}
    \label{fig:bird-gen-mask}
\end{figure*}

\begin{figure*}
    \centerfloat
    \includegraphics[width=0.55\textwidth]{cub_bg-vae_1e1-L2_1e0-crease/0011_no-shadow.jpg}
    \includegraphics[width=0.55\textwidth]{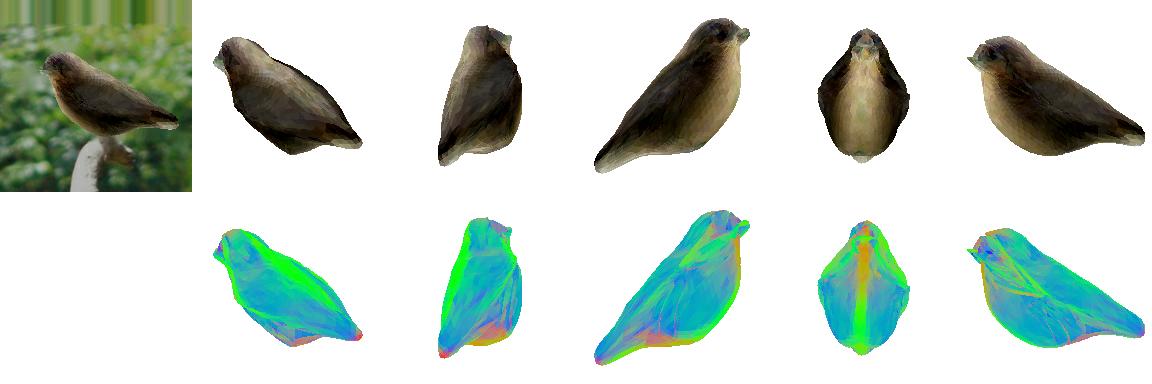} \\
    \includegraphics[width=0.55\textwidth]{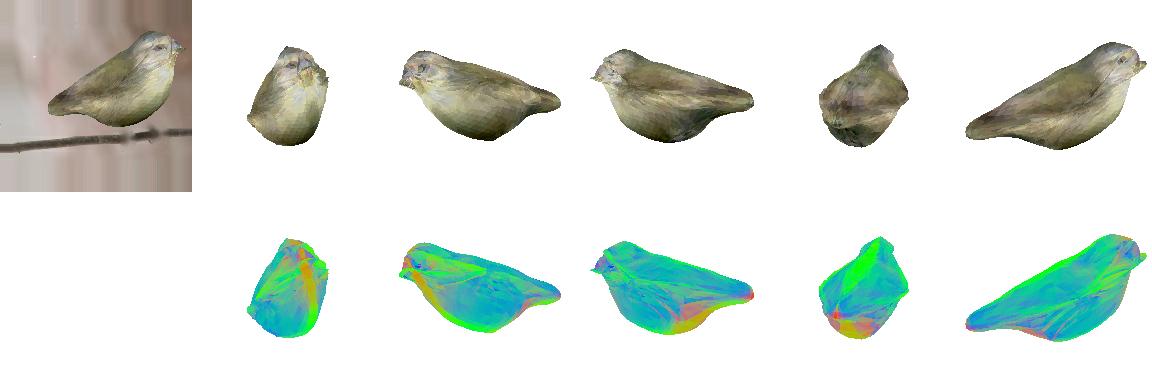}
    \includegraphics[width=0.55\textwidth]{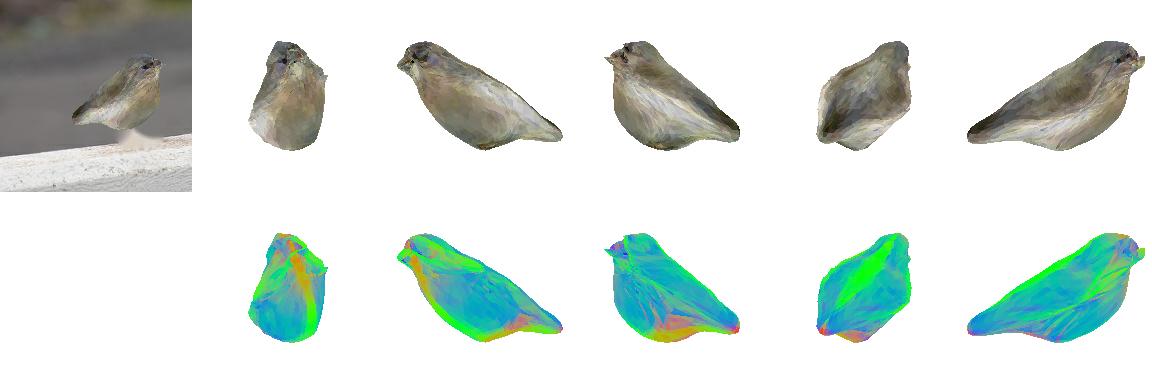} \\
    \includegraphics[width=0.55\textwidth]{cub_bg-vae_1e1-L2_1e0-crease/0015_no-shadow.jpg}
    \includegraphics[width=0.55\textwidth]{cub_bg-vae_1e1-L2_1e0-crease/0016_no-shadow.jpg} \\
    \includegraphics[width=0.55\textwidth]{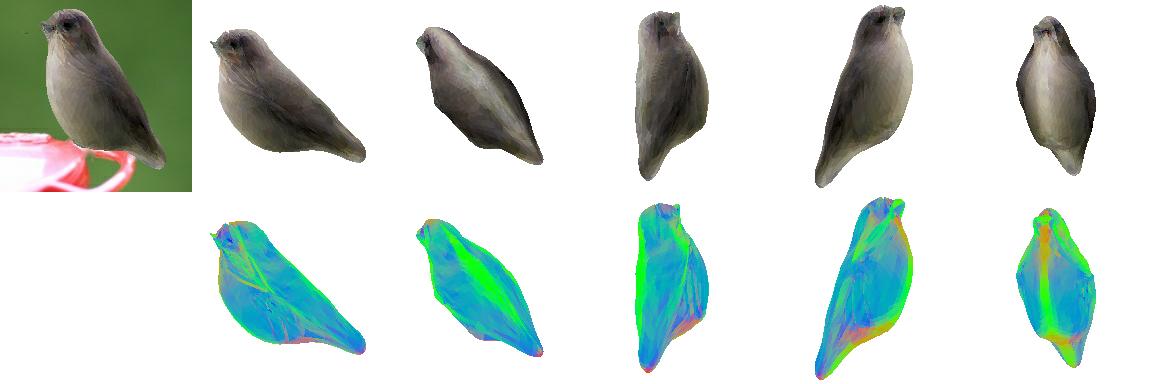}
    \includegraphics[width=0.55\textwidth]{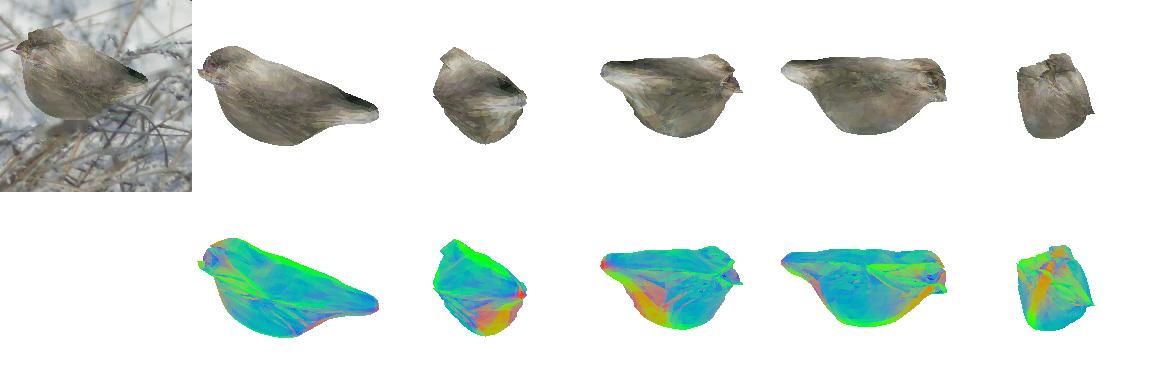} \\
    \includegraphics[width=0.55\textwidth]{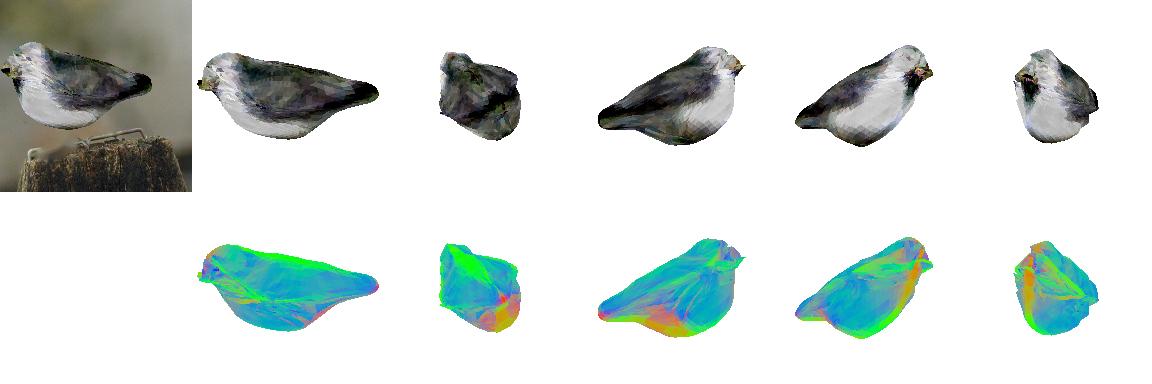}
    \includegraphics[width=0.55\textwidth]{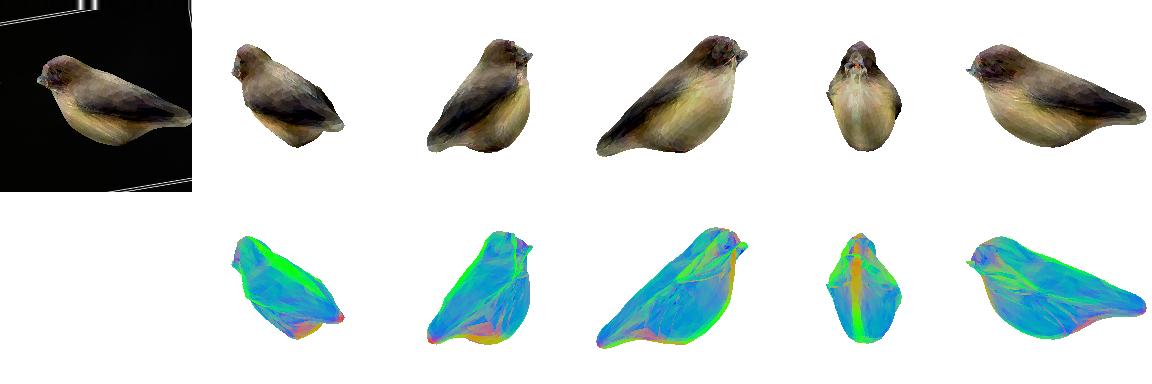} \\
    \includegraphics[width=0.55\textwidth]{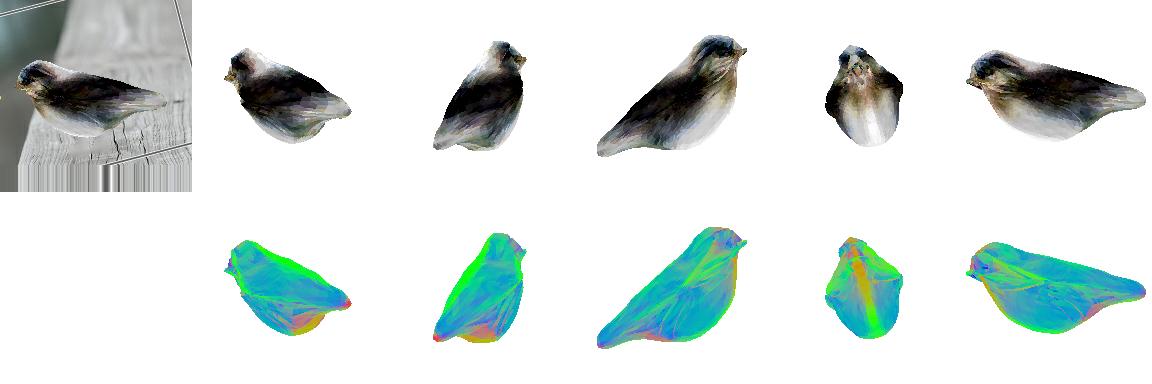}
    \includegraphics[width=0.55\textwidth]{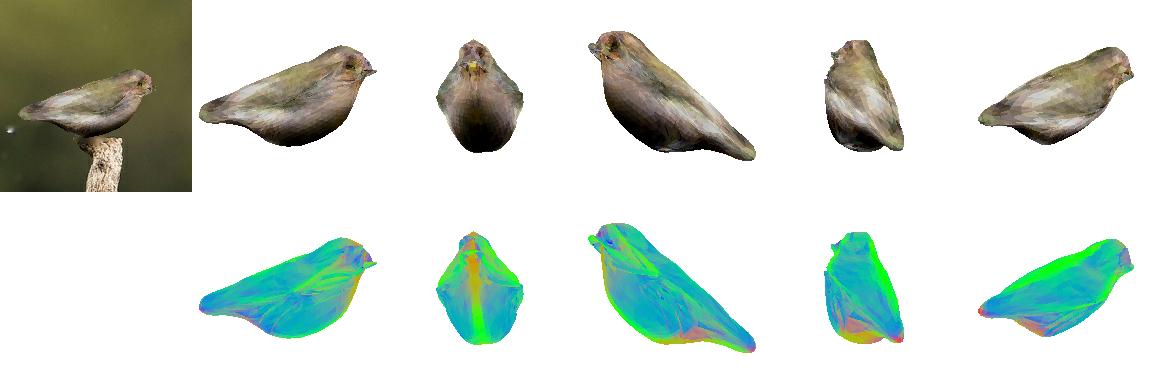}
    \caption{Examples of birds generated by our model, in setting \bgvaesetting{}.}
    \label{fig:bird-gen-bgvae}
\end{figure*}

Random (uncurated) samples from the models trained on natural images of
cars are given in \fig{car-gen-mask} and \fig{car-gen-bgvae},
and birds in \fig{bird-gen-mask} and \fig{bird-gen-bgvae}.
See \sect{experiments-real} for discussion of these results.

\begin{figure*}
    \centering
    \vspace{-1cm}
    \includegraphics[width=0.75\textwidth]{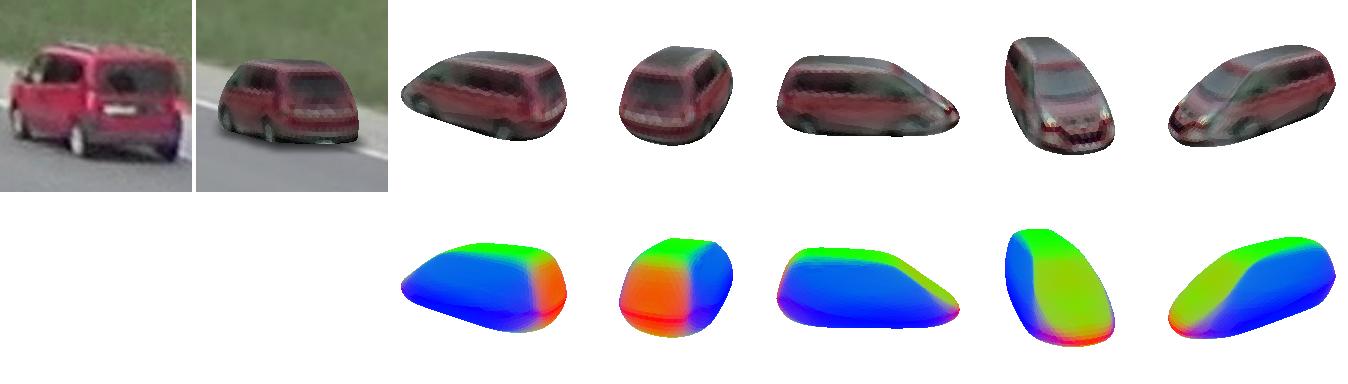} \\
    \includegraphics[width=0.75\textwidth]{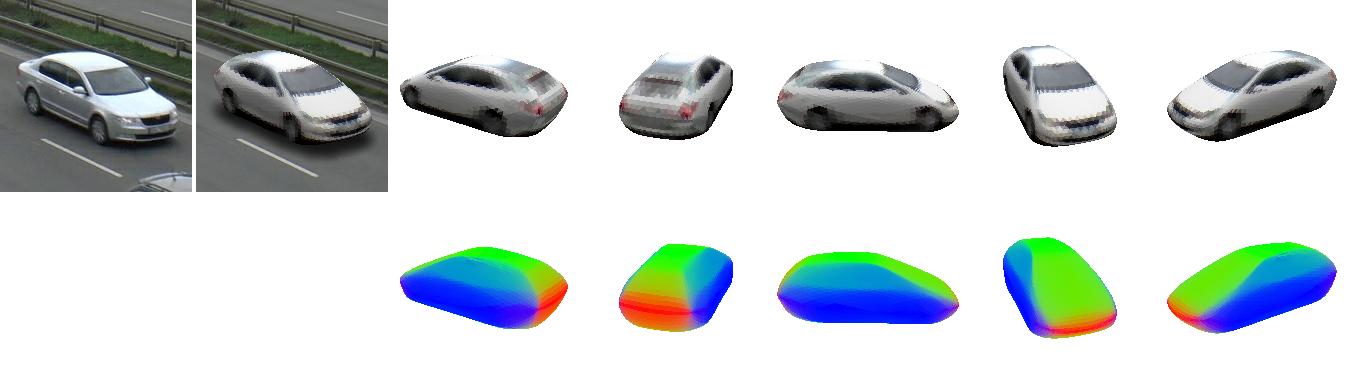} \\
    \includegraphics[width=0.75\textwidth]{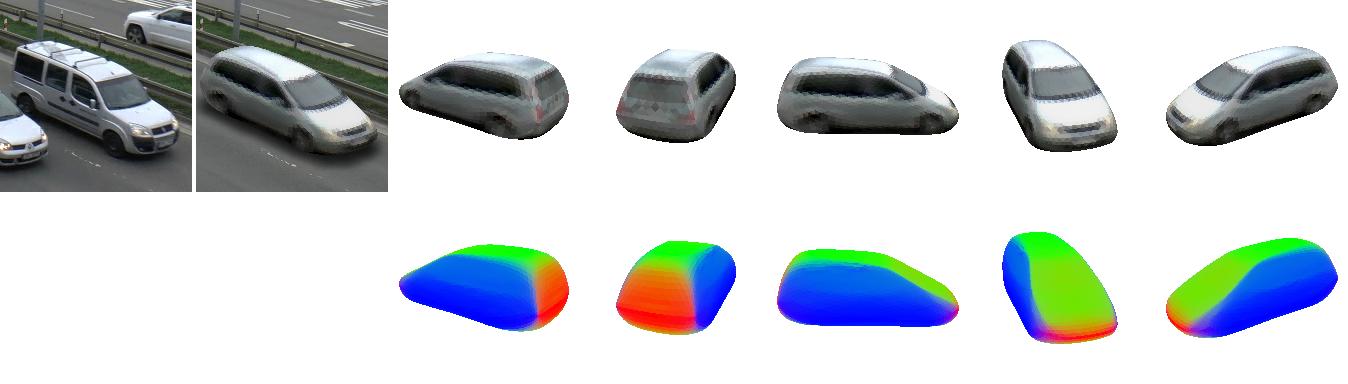}
    \includegraphics[width=0.75\textwidth]{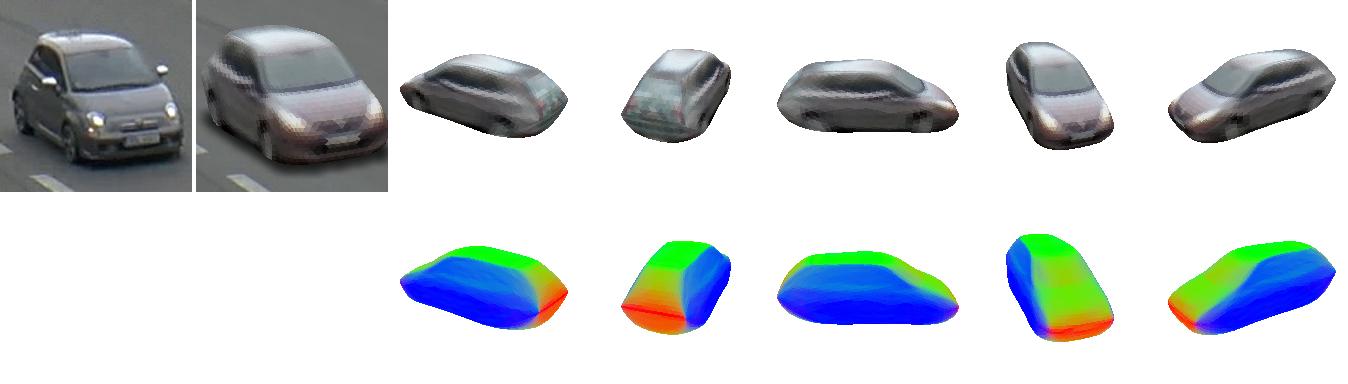} \\
    \includegraphics[width=0.75\textwidth]{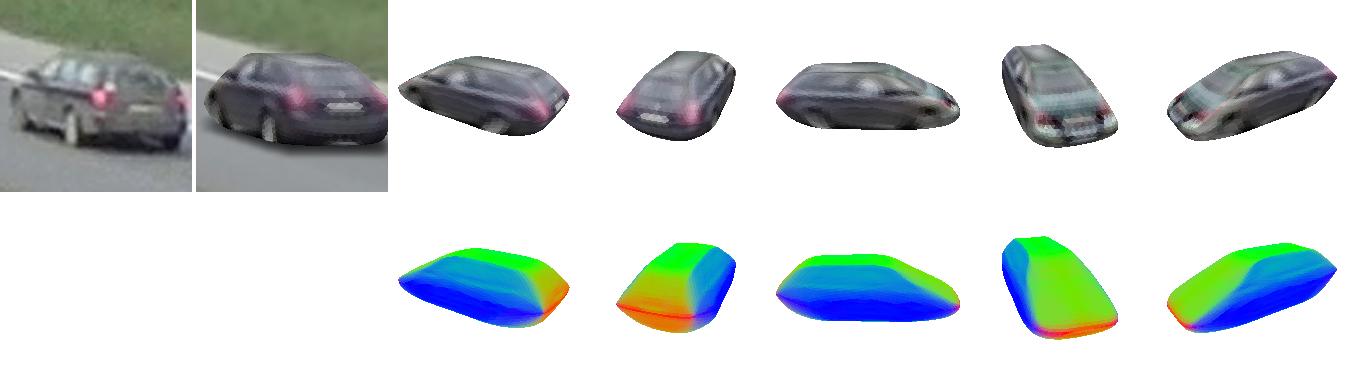} \\
    \includegraphics[width=0.75\textwidth]{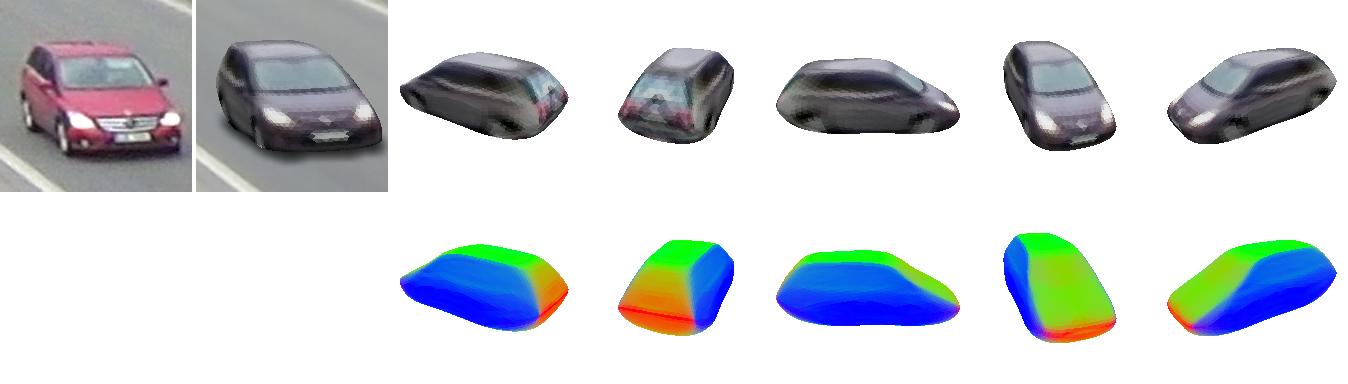}
    \caption{
        Additional examples of car reconstructions, in setting \masksetting{} (top three) and \bgvaesetting{} (bottom three).
        The left-hand image is the input to our model; the next is our reconstruction, rendered over the ground-truth
        background; the remaining five columns are different views of the reconstructed instance, with normal-maps below
    }
    \label{fig:car-recon}
\end{figure*}

\begin{figure*}
    \centering
    \vspace{-1cm}
    \includegraphics[width=0.75\textwidth]{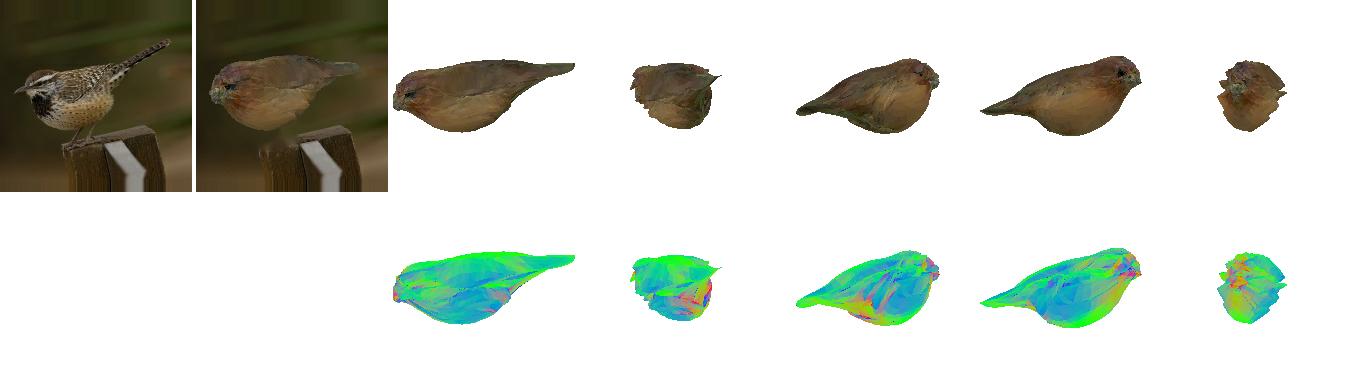} \\
    \includegraphics[width=0.75\textwidth]{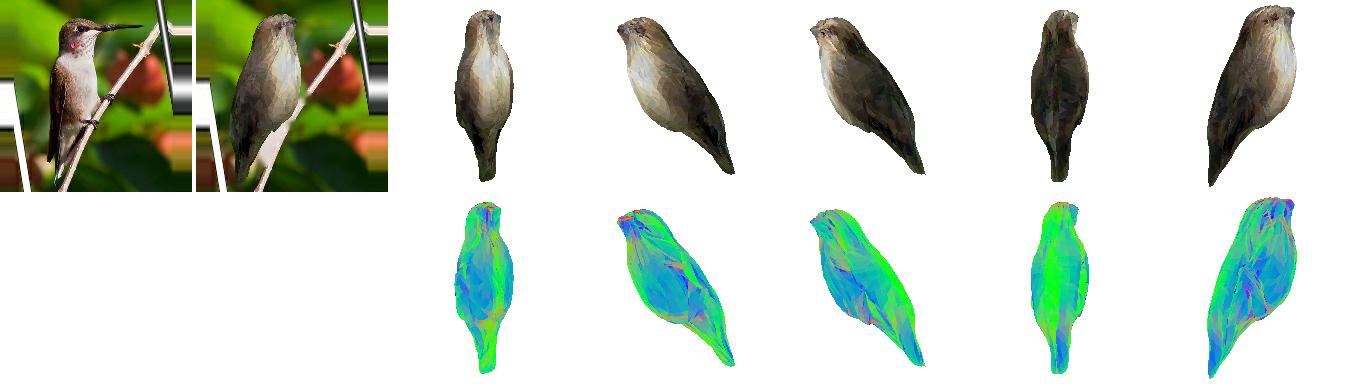} \\
    \includegraphics[width=0.75\textwidth]{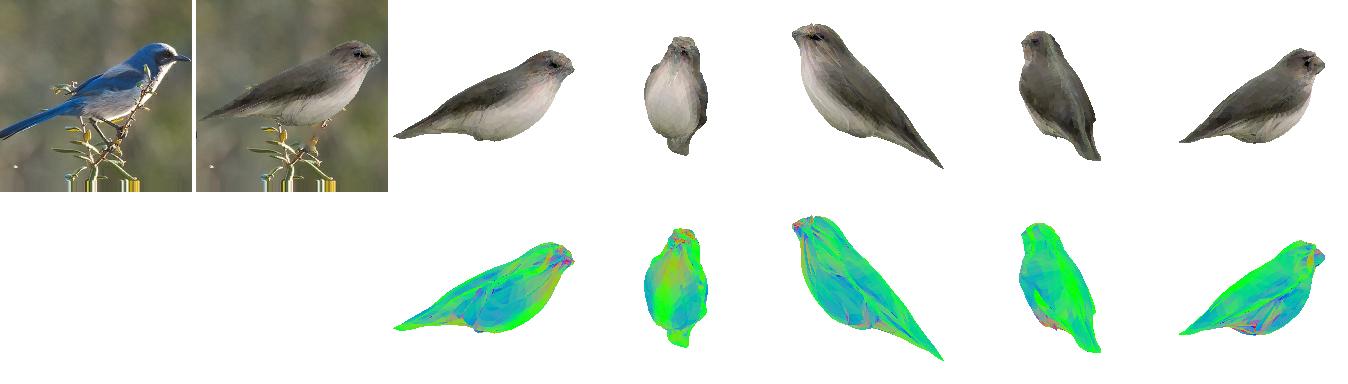}
    \includegraphics[width=0.75\textwidth]{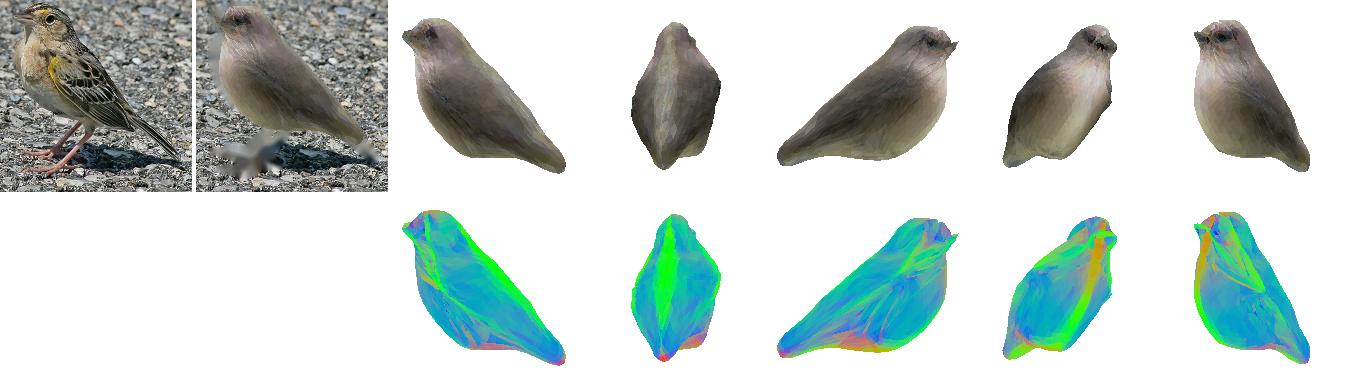} \\
    \includegraphics[width=0.75\textwidth]{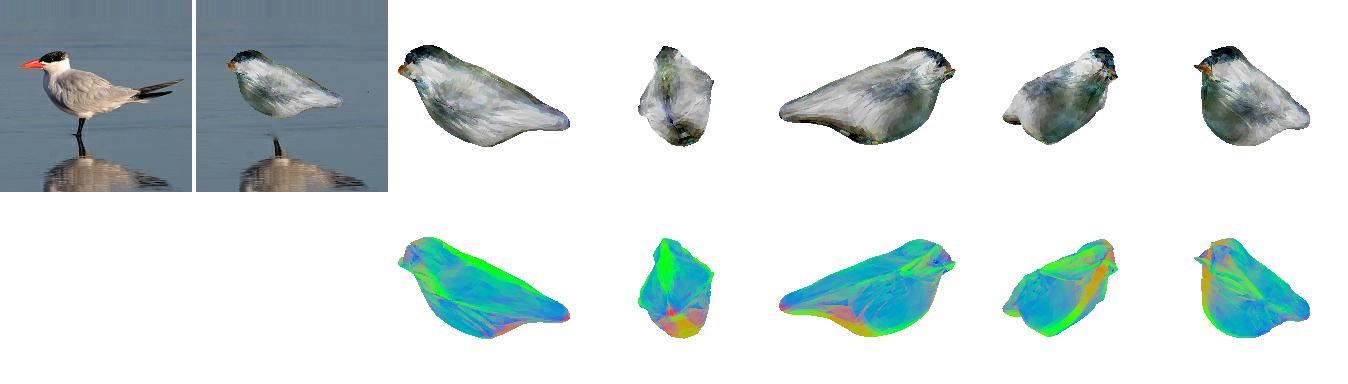} \\
    \includegraphics[width=0.75\textwidth]{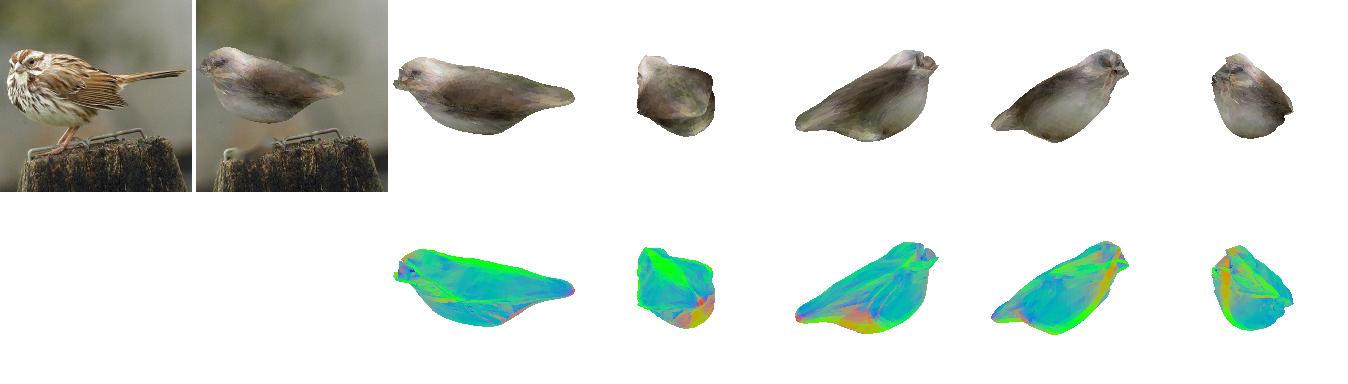} \\
    \caption{
        Additional examples of bird reconstructions, in setting \masksetting{} (top three) and \bgvaesetting{} (bottom three).
        The left-hand image is the input to our model; the next is our reconstruction, rendered over a pseudo-background (see text);
        the remaining five columns are different views of the reconstructed instance, with normal-maps below
    }
    \label{fig:bird-recon}
\end{figure*}

Additional reconstruction results on natural images of cars are given in \fig{car-recon}, and birds in \fig{bird-recon}.
See \sect{experiments-recon} for discussion of these results.

\section{Network architectures}
\label{app:architectures}

\setlist{noitemsep}

In this section, we describe the neural network architectures for each component of our model.

\paragraph{Notation.}
We use the following notation for network layers:
\begin{itemize}
\item Convolutional layers are denoted by Conv(channels,~filter~size); stride is one unless otherwise specified
\item Densely-connected layers are denoted by Dense(channels)
\item Bilinear upsampling layers are denoated by Upsampling(factor)
\item Reshape(shape) denotes reshaping the input tensor to the given shape
\item $\oplus$ denotes vector concatenation
\item $\sigma$ denotes the logistic sigmoid function
\end{itemize}
When the input to a layer is not just the output of the previous layer, we indicate this
input in a second pair of parentheses.

\paragraph{Encoders.}
The encoders consist of a shared CNN that extracts features from the input image, followed by densely-connected layers operating
on the resulting features to give predictions for each latent variable.
The feature extractor is as follows (all convolution/dense layers use relu activation and group-normalization):
\begin{itemize}
\item \textit{input: RGB image}
\item Conv$(32, 3\times3, \text{stride}=2)$
\item Conv$(64, 3\times3)$
\item MaxPool$(2\times2)$
\item Conv$(96, 3\times3)$
\item MaxPool$(2\times2)$
\item Conv$(128, 3\times3)$
\item MaxPool$(2\times2)$
\item Conv$(128, 4\times4)$
\item Dense$(128)$
\item \textit{output: 128D feature vector $\mathbf{f}$}
\end{itemize}
The shape encoder is as follows:
\begin{itemize}
\item \textit{input: 128D feature vector $\mathbf{f}$}
\item Dense$(32 \times 2)$
\item \textit{output: mean and stddev of 32D shape embedding $\mathbf{z}_\mathrm{shape}$, latter with softplus activation}
\end{itemize}
The texture encoder is as follows:
\begin{itemize}
\item \textit{input: 128D feature vector $\mathbf{f}$ and vector $\mathbf{c}$ of mean pixel colors clipping each face}
\item Dense$(96, \text{activation}=\text{relu}, \text{group-}\\\text{normalization})(\mathbf{c}) \oplus \mathbf{f}$
\item Dense$(128 \times 2)$
\item \textit{output: mean and stddev of 128D texture embedding $\mathbf{z}_\mathrm{color}$, latter with softplus activation}
\end{itemize}
The background encoder is as follows:
\begin{itemize}
\item \textit{input: 128D feature vector $\mathbf{f}$}
\item Dense$(64, \text{activation}=\text{relu}, \text{group-normalization})$
\item Dense$(16 \times 2)$
\item \textit{output: mean and stddev of 16D background embedding $\mathbf{z}_\mathrm{bg}$, latter with softplus activation}
\end{itemize}
The pose encoder is as follows:
\begin{itemize}
\item \textit{input: 128D feature vector $\mathbf{f}$}
\item Dense$(5)$
\item \textit{output: 2D offset in xz-plane; 3D log-scale}
\end{itemize}

\paragraph{Decoders.}
The decoders consist of densely-connected networks, taking the latent variables as input.
The shape decoder $\mathrm{dec}_\mathrm{shape}$ is as follows:
\begin{itemize}
\item \textit{input: 32D shape embedding $\mathbf{z}_\mathrm{shape}$}
\item Dense$(32, \text{activation}=\text{elu})$
\item Dense$(3 N_V)$
\item Reshape$(N_V, 3)$
\item \textit{output: 3D offset vectors to be added to each of the $N_V$ vertices of the base mesh}
\end{itemize}
The texture decoder $\mathrm{dec}_\mathrm{color}$ is as follows:
\begin{itemize}
\item \textit{input: 128D texture embedding $\mathbf{z}_\mathrm{color}$ and 32D shape embedding $\mathbf{z}_\mathrm{shape}$}
\item Dense$(128, \text{activation}=\text{elu})(\mathbf{z}_\mathrm{color} \oplus \mathbf{z}_\mathrm{shape}) + \mathbf{z}_\mathrm{color}$
\item Dense$(192, \text{activation}=\text{elu})$
\item Dense$(3 N_F)$
\item Reshape$(N_F, 3) \, / \, 10 + \frac{1}{2}$
\item \textit{output: RGB colors for each of the $N_F$ faces of the mesh}
\end{itemize}
The background decoder, used only in setting \bgvaesetting{}, is as follows (all convolution layers use elu activation, except the last):
\begin{itemize}
\item \textit{input: 16D background embedding}
\item Reshape$(1\times1\times16)$
\item Upsample$(4\times)$
\item Conv$(64, 3\times3)$
\item Upsample$(2\times)$
\item Conv$(32, 3\times3)$
\item Upsample$(2\times)$
\item Conv$(16, 3\times3)$
\item Upsample$(2\times)$
\item $\sigma(\text{Conv}(3, 3\times3) \, / \, 2$)
\item Upsample$(4\times$ or $6\times)$
\item \textit{output: RGB background image}
\end{itemize}

\paragraph{Baseline VAE.}
The encoder for the baseline VAE uses the same feature extractor as the main model.
To convert the features to the mean and variance of the latent variables, the following architecture is used:
\begin{itemize}
\item \textit{input: 128D feature vector $\mathbf{f}$}
\item Dense$(160 \times 2)$
\item \textit{output: mean and stddev of 160D image embedding, latter with softplus activation}
\end{itemize}
The decoder is as follows (all layers use elu activation, except the last):
\begin{itemize}
\item \textit{input: 160D image embedding}
\item Dense$(256)$
\item Reshape$(4\times4\times16)$
\item Conv$(128, 3\times3)$
\item Upsample$(2\times)$
\item Conv$(64, 3\times3)$
\item Upsample$(2\times)$
\item Conv$(32, 3\times3)$
\item Upsample$(2\times)$
\item Conv$(24, 3\times3)$
\item Upsample$(2\times)$
\item Conv$(16, 3\times3)$
\item Upsample$(2\times)$
\item Conv$(3, 3\times3) + \frac{1}{2}$
\item \textit{output: RGB image}
\end{itemize}

\newcommand{\shgenfig}[3]{
\begin{figure*}
    \centerfloat
    \vspace{-1cm}
    \includegraphics[width=0.55\textwidth]{#3/0000_no-shadow.jpg}
    \includegraphics[width=0.55\textwidth]{#3/0001_no-shadow.jpg} \\
    \includegraphics[width=0.55\textwidth]{#3/0002_no-shadow.jpg}
    \includegraphics[width=0.55\textwidth]{#3/0003_no-shadow.jpg} \\
    \includegraphics[width=0.55\textwidth]{#3/0004_no-shadow.jpg}
    \includegraphics[width=0.55\textwidth]{#3/0005_no-shadow.jpg} \\
    \includegraphics[width=0.55\textwidth]{#3/0006_no-shadow.jpg}
    \includegraphics[width=0.55\textwidth]{#3/0007_no-shadow.jpg} \\
    \includegraphics[width=0.55\textwidth]{#3/0008_no-shadow.jpg}
    \includegraphics[width=0.55\textwidth]{#3/0009_no-shadow.jpg} \\
    \includegraphics[width=0.55\textwidth]{#3/0010_no-shadow.jpg}
    \includegraphics[width=0.55\textwidth]{#3/0011_no-shadow.jpg}
    \caption{#1}
    \label{fig:#2}
\end{figure*}
}

\shgenfig{Examples of cars generated by our model, in setting \masksetting{} with parametrization \densep{}.}{sh-car-gen-mask}{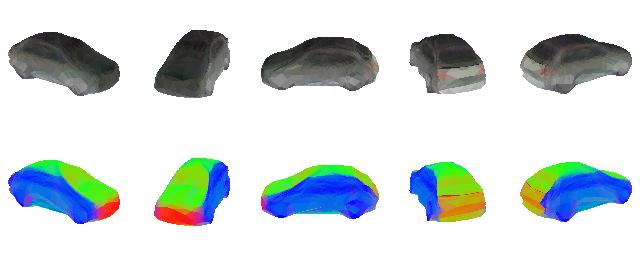}
\shgenfig{Examples of cars generated by our model, in setting \bgvaesetting{} with parametrization \densep{}.}{sh-car-gen-bgvae}{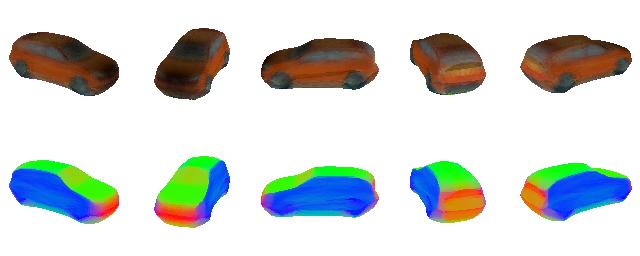}

\shgenfig{Examples of chairs generated by our model, in setting \masksetting{} with parametrization \densep{}.}{sh-chair-gen-mask-dense}{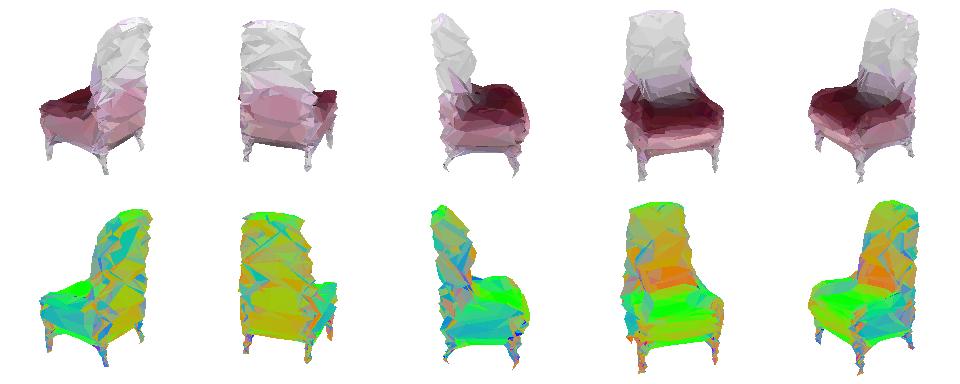}
\shgenfig{Examples of chairs generated by our model, in setting \masksetting{} with parametrization \nimp{}.}{sh-chair-gen-mask-nimp}{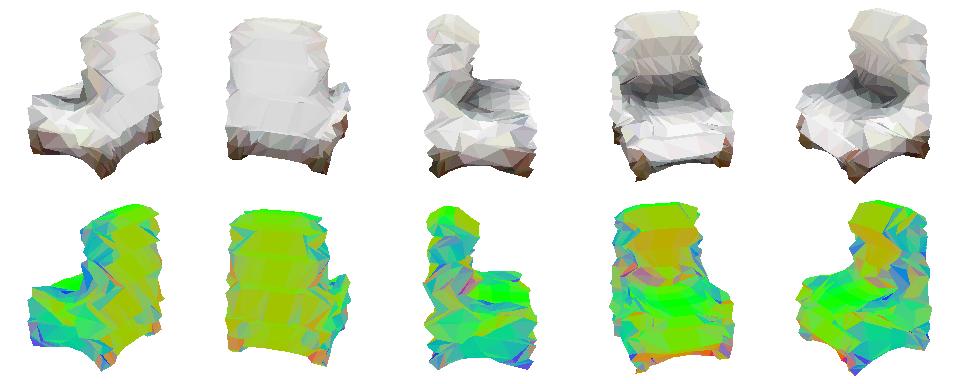}
\shgenfig{Examples of chairs generated by our model, in setting \bgvaesetting{} with parametrization \densep{}.}{sh-chair-gen-bgvae}{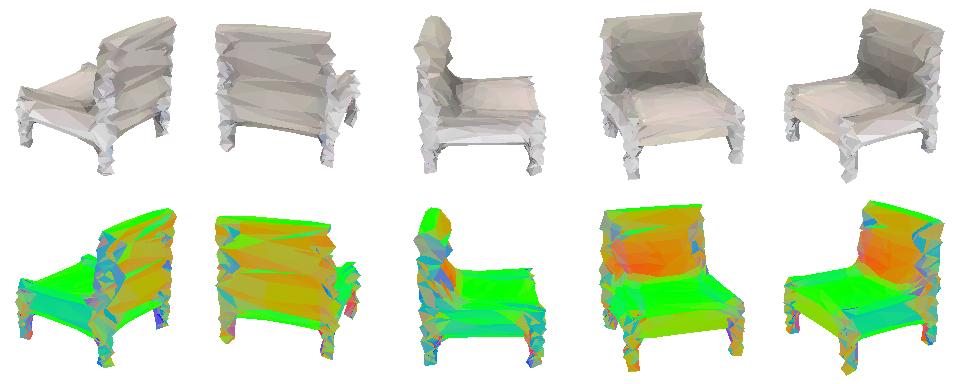}

\shgenfig{Examples of airplanes generated by our model, in setting \masksetting{} with parametrization \densep{}.}{sh-airplane-gen-mask-dense}{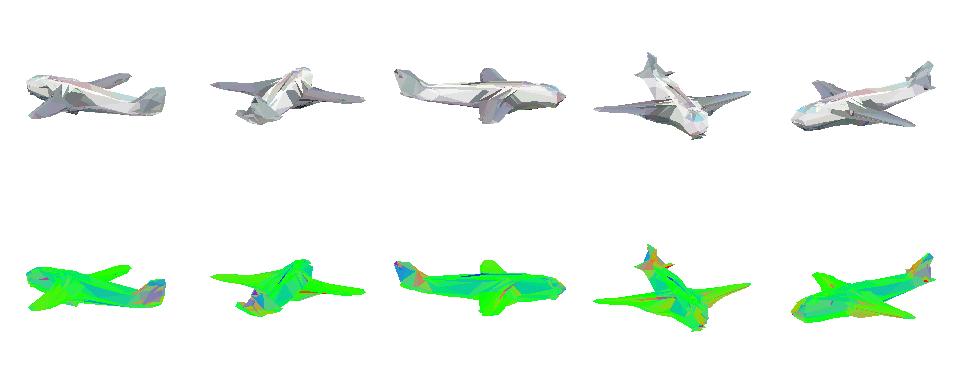}
\shgenfig{Examples of airplanes generated by our model, in setting \masksetting{} with parametrization \nimp{}.}{sh-airplane-gen-mask-nimp}{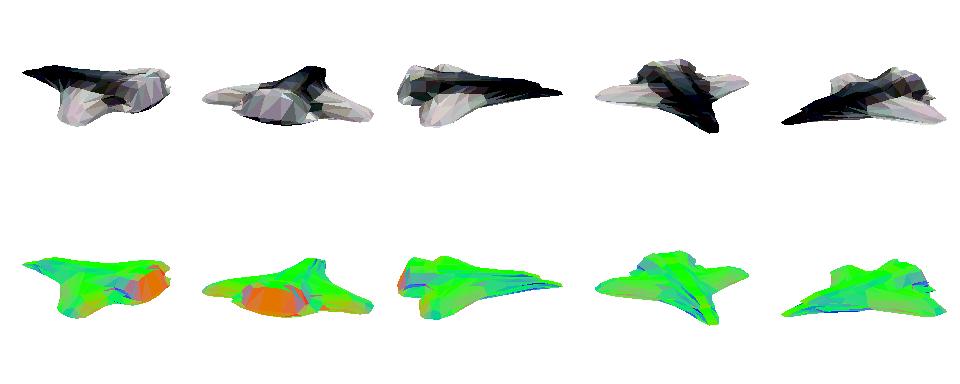}
\shgenfig{Examples of airplanes generated by our model, in setting \bgvaesetting{} with parametrization \densep{}.}{sh-airplane-gen-bgvae}{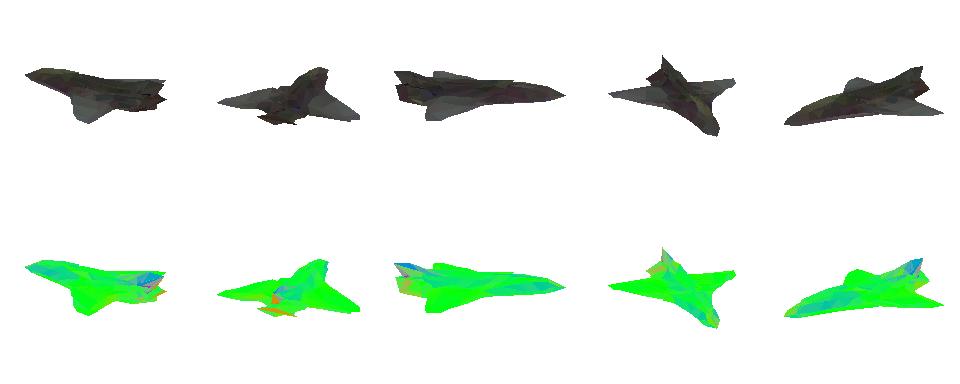}

\shgenfig{Examples of sofas generated by our model, in setting \masksetting{} with parametrization \densep{}.}{sh-sofa-gen-mask-dense}{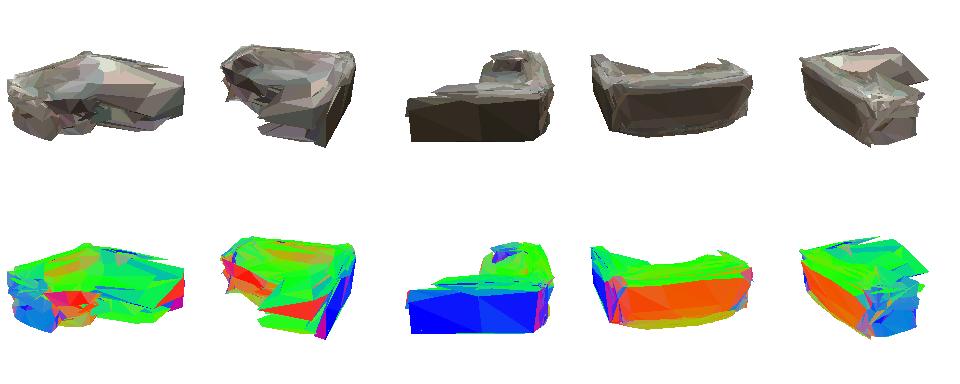}
\shgenfig{Examples of sofas generated by our model, in setting \masksetting{} with parametrization \nimp{}.}{sh-sofa-gen-mask-nimp}{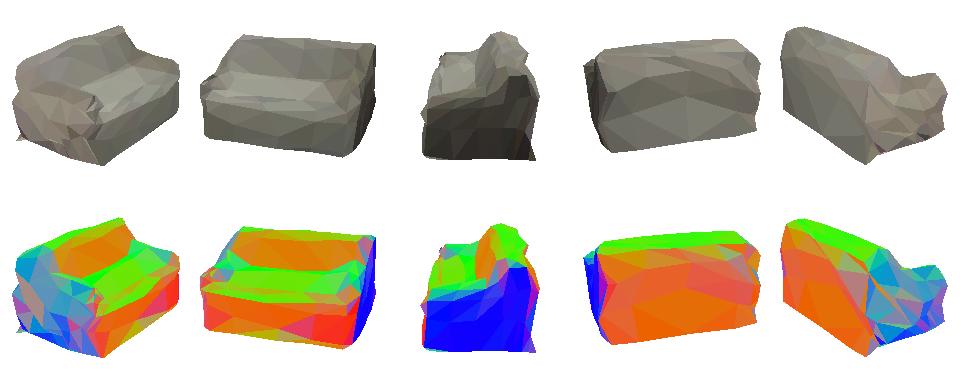}
\shgenfig{Examples of sofas generated by our model, in setting \bgvaesetting{} with parametrization \densep{}.}{sh-sofa-gen-bgvae}{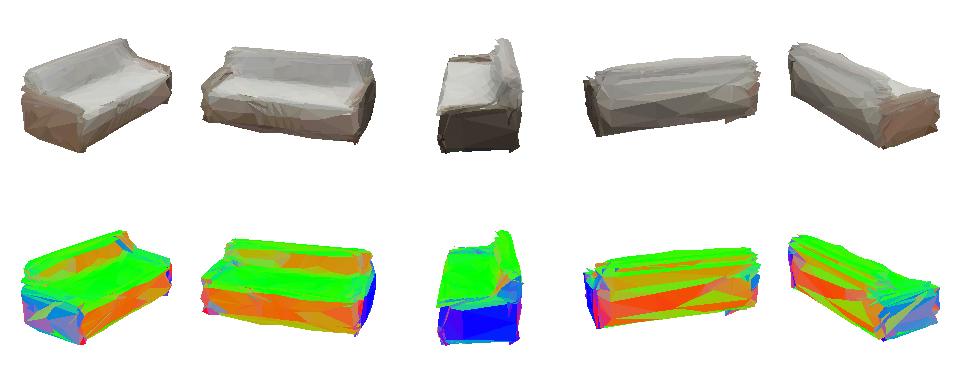}

\end{document}